\documentclass[10pt,journal,compsoc]{IEEEtran}
\usepackage{graphicx}
\usepackage{amsmath}
\usepackage{makecell}
\usepackage{mathrsfs}
\usepackage{gensymb}
\usepackage[table]{xcolor}
\usepackage{footnote}
\usepackage{amssymb}
\usepackage{multirow}
\usepackage{diagbox}
\usepackage{subcaption}
\usepackage{stfloats}

\begin{document}
	
	\title{RealGait: Gait Recognition for Person Re-Identification}

	%
	%
	%
	%
	
	\author{Shaoxiong~Zhang,
		Yunhong~Wang,~\IEEEmembership{Fellow,~IEEE,}
		Tianrui~Chai,
		Annan~Li,~\IEEEmembership{Member,~IEEE,}
		and~Anil~K.~Jain,~\IEEEmembership{Life~Fellow,~IEEE}%

		
		\IEEEcompsocitemizethanks{\IEEEcompsocthanksitem Shaoxiong~Zhang, Yunhong~Wang, Tianrui~Chai, and Annan~Li are with the School of Computer Science and Engineering, Beihang University, Beijing, China.
			E-mail: \{zhangsx, yhwang, trchai, liannan\}@buaa.edu.cn

			\IEEEcompsocthanksitem Anil K. Jain is with the Department of Computer Science and Engineering, Michigan State University, East Lansing, MI 48824 USA.
			E-mail: jain@cse.msu.edu
			
			\IEEEcompsocthanksitem Corresponding author: Annan~Li, Yunhong~Wang.}

	}
	
	\markboth{}%
	{Zhang \MakeLowercase{\textit{et al.}}:RealGait: Gait Recognition for Person Re-Identification:}
	
	\IEEEtitleabstractindextext{%
		\begin{abstract}
			
			Human gait is considered a unique biometric identifier which can be acquired in a covert manner at a distance.
			However, models trained on existing public domain gait datasets which are captured in controlled scenarios lead to drastic performance decline when applied to real-world unconstrained gait data.
			On the other hand, video person re-identification techniques have achieved promising performance on large-scale publicly available datasets.
			Given the diversity of clothing characteristics, clothing cue is not reliable for person recognition in general. 
			So, it is actually not clear why the state-of-the-art person re-identification methods work as well as they do.
			In this paper, we construct a new gait dataset by extracting silhouettes from an existing video person re-identification challenge which consists of 1,404 persons walking in an unconstrained manner.
			Based on this dataset, a consistent and comparative study between gait recognition and person re-identification can be carried out.
			Given that our experimental results show that current gait recognition approaches designed under data collected in controlled scenarios are inappropriate for real surveillance scenarios, we propose a novel gait recognition method, called RealGait.
			%
			%
			Our results suggest that recognizing people by their gait in real surveillance scenarios is feasible and the underlying gait pattern is probably the true reason why video person re-idenfification works in practice.
			
		\end{abstract}
		
		}

	\maketitle
	
	\IEEEdisplaynontitleabstractindextext
	
	\IEEEpeerreviewmaketitle

	\IEEEraisesectionheading{
		\section{Introduction}
		\label{sec:intro}}
	
	\IEEEPARstart{G}{ait} recognition is a biometric technology that identifies people by their body shape and the way they walk. Different from other biometrics like face, fingerprint, or iris, gait can be captured at a distance using off the shelf sensors in a covert manner. For these reasons, gait is a promising biometric trait for scenarios where face cannot be observed with sufficient resolution for recognition. Prior art has demonstrated the feasibility of gait recognition in specific tasks, for example, age and gender estimation~\cite{zhang2019gait,Xu2017}, identification with carried objects~\cite{Li2020b} and cross-view identification~\cite{Song.2019,Zhang2019c,Chao11}.
	
	Several many deep learning models~\cite{Chao11, Zhang2019b, Fan2020} have also been proposed for gait recognition which attain excellent performance on large gait datasets with over 10,000 subjects~\cite{Takemura2018}. Although this progress is encouraging, gait recognition still suffers from a major problem: the scenario for capturing existing gait datasets differs greatly from the real-world applications. For example, in CASIA-B dataset~\cite{Yu2006a} and OU-MVLP dataset~\cite{Takemura2018}, the two most popular gait datasets, the person's silhouette is obtained either by subtracting simple static background from the image or using the chroma key compositing. More importantly, in such settings, subjects are asked to walk in a straight-line path, and there are no other people in the camera's field of view. Such stringent requirements are not feasible in person re-identification or surveillance scenarios.
	
	The above limitation has already been noticed in the literature. To better simulate the real scenario, a few gait datasets, i.e. the FVG dataset~\cite{Zhang2019d} and SMVDU-Multi-Gait dataset \cite{Singh2019}, have been collected in outdoor environments. Although illumination is natural in such datasets, the field of view contains a single subject that follows designated routes. In real-world scenarios, there are probably multiple subjects who walk in arbitrary directions. These restrictions in publicly available datasets severely limit the application scenarios of gait recognition techniques trained on constrained datasets.
	
	\begin{figure}[t]
		\centering
		\includegraphics[width=1\linewidth]{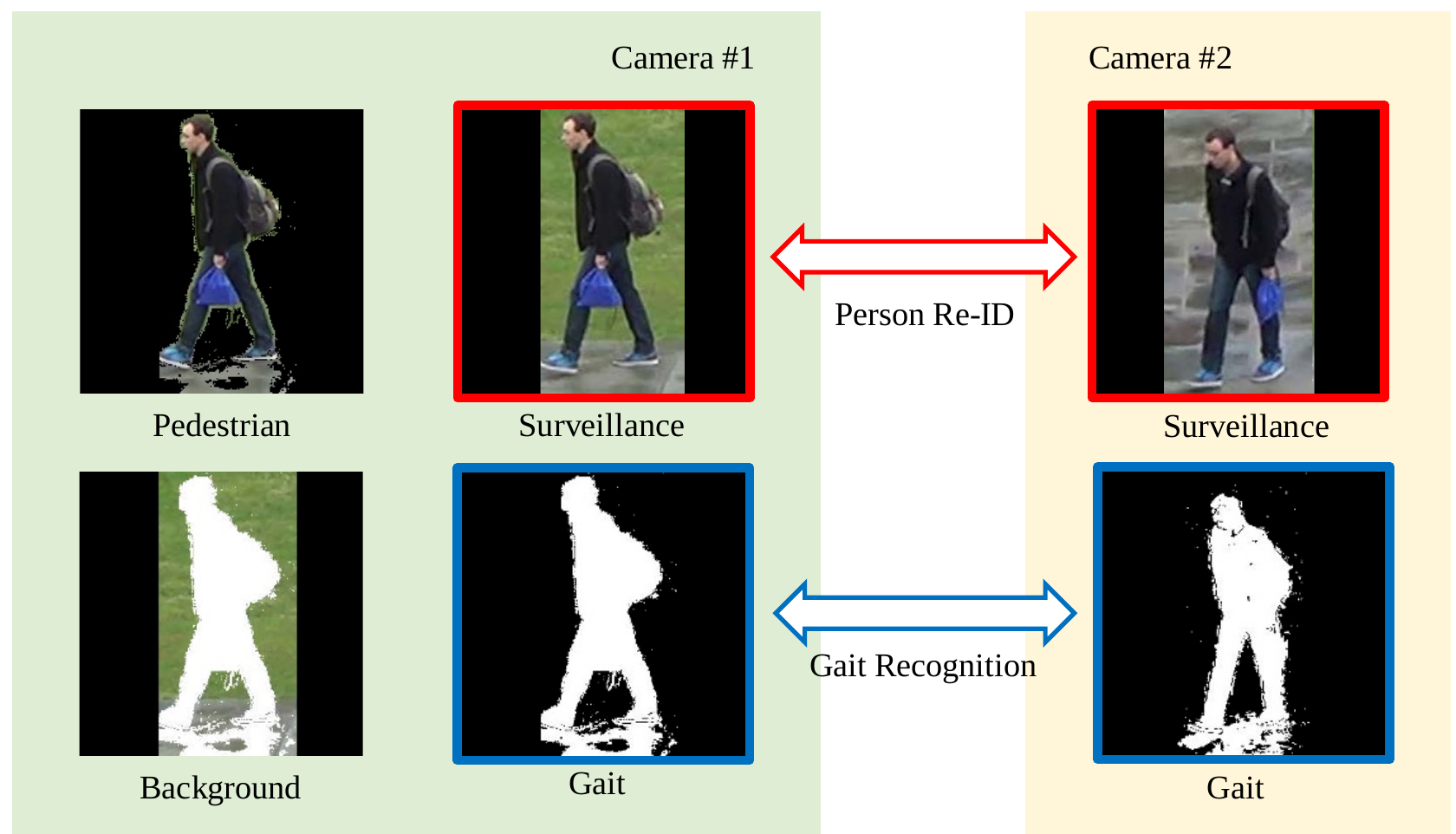}
		\vspace{-3mm}
		\caption{Person re-identification vs. gait recognition in the wild.}
		\label{fig:intro}
	\end{figure}
	
	Another pedestrian recognition task, i.e. person re-identification (Re-ID)~\cite{Yao2019} has attracted increasing attention (see Figure~\ref{fig:intro}). The original person Re-ID task involved association of a limited number of people across multiple cameras with a non-overlapping field of views within a restricted space and duration. In this case, clothing diversity can be a reliable cue for identifying people. However, with the advances in person Re-ID techniques, the task has been extended to general large-scale pedestrian recognition tasks in outdoor scenarios. Although promising performance has been achieved for person Re-ID in a general setting, the diversity of clothing characteristics is no longer a reliable cue for recognition. Some studies have even investigated the case of subjects wearing different clothing~\cite{pami021clothes_change}. In other words, the reason why extended person Re-ID can achieve high accuracy is not clear\footnote{In this paper, the person re-identification is actually the video person re-identification. Since no motion cue is available, the static image-based person re-identification cannot be compared with gait recognition.}.
	
	This work is inspired by the following observations. To address the problem of gait recognition in arbitrary settings or ``in the wild'', (i) real surveillance videos need to be utilized, and (ii) gait biometric trait is a major cue for large-scale person Re-ID. As illustrated in Figure~\ref{fig:intro}, the main difference between person identification or re-identification and gait recognition in the wild is simply the clothing color appearance. Since the clothing characteristics are not considered reliable for person recognition, gait is probably the real reason why current algorithms for person Re-ID provide good recognition accuracy. To address the above observations, we pose the following two questions. (i) How does gait contribute to person recognition in surveillance videos? (ii) What is the impact of gait features in person Re-ID? When these two questions are put together, they indeed are two sides of the same coin.
	
	To address the role of gait in person Re-ID, we construct a new gait dataset, named \emph{BUAA-Duke-Gait} dataset by extracting silhouettes from an existing video person Re-ID dataset, \emph{DukeMTMC-VideoReID}~\cite{wu2018cvpr_oneshot}. We believe this dataset will serve as a new challenge for gait recognition. In addition, this dataset will allow a strict and consistent comparative study between gait recognition and general person identification or Re-ID. To make this feasible, we first develop an effective method of extracting gait silhouettes from surveillance videos. Then a novel approach for gait recognition in the wild is proposed. Experimental results show that gait recognition in surveillance videos is significantly more challenging than gait recognition in controlled video capture. The proposed gait recognition method achieves 78.24\% rank-1 accuracy using the binary silhouettes alone, while the corresponding rank-1 accuracy of full color version, i.e. the person re-identification, is 94\%. This demonstrates that gait indeed plays a key role in person re-identification. 
	
	In summary, we make the following major contributions in this paper.
	\begin{itemize}
		\item We present a large-scale gait dataset named \emph{BUAA-Duke-Gait} derived from an existing video person re-identification dataset\footnote{The \emph{BUAA-Duke-Gait} dataset will be made publicly available after this paper is accepted for publication.}. This dataset not only poses a new challenge for gait recognition, to our knowledge, it is also the first gait benchmark that can be directly evaluated for person re-identification tasks.
		\item We investigate the performance of three state-of-the-art gait recognition models on the new dataset, \textit{BUAA-Duke-Gait}. Results show that gait models developed for data captured in controlled environments cannot be adopted for either gait recognition or ReID in surveillance data.
		\item We propose a deep network model, \emph{RealGait}, for cross-scene gait recognition, which sets a new state-of-the-art for gait recognition performance in the wild.
		\item Extensive experiments are also conducted to elicit the impact of gait biometric in general video person re-identification.
	\end{itemize}
	
	The remainder of this paper is organized as follows. In Section~\ref{sec:related}, we first review relevant studies. Details of the proposed new dataset are presented in Section~\ref{sec:data}. Section~\ref{sec:method} describes the proposed new gait recognition method, while the experimental results are given in Section~\ref{sec:exp}. Finally, we conclude in Section~\ref{sec:conclude}.
	
	\section{Related Work}
	\label{sec:related}
	
	In this section, we first review the development of gait datasets. Since deep learning-based gait recognition can be roughly grouped into two categories, i.e. appearance-based and model-based methods, we introduce the literature on gait recognition accordingly. Because of relevance to gait recognition, a brief review of video person re-identification is also presented. 
	
	\subsection{Gait Datasets}
	A summary of known public gait datasets is shown in Table~\ref{tab:datasets}. To facilitate the extraction of pedestrian silhouettes, most gait images are captured using static cameras against a simple background or even in front of a green screen. To further control the ambient illumination, some datasets are captured in indoor environments. Such restrictions limit the study of gait recognition in real-world scenarios. More importantly, subjects in these datasets are cooperative and they are asked to walk in a straight trajectory towards the camera or the center of a camera array. Consequently, the angle between the camera and the pedestrian walking direction is fixed. The coverage of viewpoint depends on the number of cameras, which is usually less than 15~\cite{Yu2006a,an2020performance}. Later on, we will show empirically that such an image capture method limits the applicability of existing gait recognition models. 
	
	In 2021, several new datasets containing gait data captured in uncontrolled scenarios were released. Mu et al.~\cite{mu2021resgait} released the ReSGait dataset, collected over a 15-month time span, where the subjects walked freely in an indoor corridor. These two characteristics (large time span and free walking pattern) have never been included in the previous gait datasets. Zhu et al.~\cite{zhu2021gait} published GREW dataset, which is a large-scale gait dataset captured in real-world scenarios. Although there are only a few gait models trained on these two gait datasets that have been published, the core research in gait recognition has started to move from controlled scenarios to real-world scenarios.
	
	\begin{table*}[t]
		\setlength{\abovecaptionskip}{1pt}
		\caption{
			Summary of existing public gait image datasets. \textit{\#Subjects} and \textit{\#Views} refer to the number of identities and the number of cameras, respectively. \textit{Environment} describes the environmental conditions during data collection, where \textit{Chroma key} means that the images are captured with a green screen background, allowing chroma keying to extract the silhouettes of subjects. \textit{Static} and \textit{Dynamic} describe if there is any other object, i.e. pedestrians and vehicles, moving in the camera's field of view as distractors. \textit{GEI} and \textit{OF} are abbreviations of gait energy image and optical flow, respectively. \textit{Sil.} refers to silhouette.}
		\renewcommand{\arraystretch}{1.6}
		\rowcolors{2}{gray!20}{white}
		\scalebox{0.77}
		{\parbox{\linewidth}{
				\begin{tabular}{ccccccccl}
					\hline
					\rowcolor{gray!50} \rule{0pt}{14pt}
					\textbf{Year} & \textbf{Dataset} & \textbf{\#Subjects} & \textbf{\#Views} & \textbf{Environment} & 
					\textbf{Multi-person} & \makecell[c]{\textbf{User}\\ \textbf{cooperation}} & \textbf{Data modalities} & \textbf{Challenges}\\ 
					2001 & CMU MoBo~\cite{gross2001cmu} & 25 & 6 & Indoor, Static & No & Yes & RGB, Sil. & Treadmill, Speeds, Carrying, Surface\\ 
					2001 & CASIA-A~\cite{wang2003silhouette} & 20 & 3 & Outdoor, Static  & No & Yes  & RGB, Sil. & Viewpoint\\ 
					2002 & SOTON~\cite{shutler2004large} & 115 & 2 & Indoor, Chroma key & No & Yes  & RGB, Sil. & Treadmill\\ 
					2005 & USF~\cite{sarkar2005humanid} & 122 & 2 & Outdoor, Static & No & Yes  & RGB & Surface, Shoes, Carrying, Duration\\ 
					2006 & CASIA-B~\cite{Yu2006a} & 124 & 11 & Indoor, Static & No & Yes  & RGB, Sil. & Viewpoint, Clothing, Carrying\\ 
					2007 & CASIA-C~\cite{tan2007uniprojective} & 153 & 1 & Outdoor, Static, Night$^a$ & No & Yes  & Infrared, Sil. & Speed, Carrying\\ 
					2012 & OU-Treadmill-A~\cite{tsuji2010silhouette} & 34 & 4 & Indoor, Chroma key & No & Yes  & Sil. & Treadmill, Speed\\ 
					2012 & OU-Treadmill-B~\cite{hossain2010clothing} & 68 & 4 & Indoor, Chroma key & No & Yes  & Sil. & Treadmill, Clothing\\ 
					2012 & OULP~\cite{iwama2012isir} & 4,016 & 4 & Indoor, Chroma key & No & Yes  & Sil. & Viewpoint\\
					2012 & TUM GAID~\cite{hofmann2014tum} & 305 & 1 & Indoor, Static & No & Yes  & RGB & Duration, Carrying, Shoes\\ 
					2017 & OULP-Age~\cite{xu2017isir} & 63,846 & 1 & Indoor, Chroma key & No & Yes  & GEI & Age\\ 
					2018 & OU-LP-Bag~\cite{uddin2018isir} & 6,2528 & 1 & Indoor, Chroma key & No & Yes  & Sil. & Carrying\\ 
					2018 & OU-MVLP~\cite{Takemura2018} & 10,307 & 14 & Indoor, Chroma key & No & Yes  & Sil. & Viewpoint\\ 
					2019 & SMVDU-Multi-Gait~\cite{Singh2019} & 20 & 3 & Outdoor, Static  & No & Yes  & RGB & Occlusion\\ 
					2019 & FVG~\cite{Zhang2019d} & 226 & 3 & Outdoor, Static & No & Yes  & RGB & Speed, Carrying, Clothing, Duration\\ 
					2019 & Outdoor-Gait~\cite{Song.2019} & 138 & 1 & Outdoor, Dynamic & No & Yes  & RGB & Carrying, Clothing\\ 
					2020 & OUMVLP-Pose~\cite{an2020performance} & 10,307 & 14 & Indoor, Chroma key & No & Yes  & Skeleton & Viewpoint\\ 
					2021 & VersatileGait~\cite{dou2021versatilegait} & 11,000 & 33 & Synthetic & Yes & N.A.  & Sil. & Viewpoint, Carrying, Clothing\\ 	
					2021 & ReSGait~\cite{mu2021resgait} & 172 & 1 & Indoor, Static & Yes & No & Sil., Pose & Free walking, Carrying, Clothing, Phone  \\
					2021 & GREW~\cite{zhu2021gait} & 26,345 & Multiple$^b$ & Outdoor, Dynamic & Yes & No & Sil., GEI, Pose, OF & Free walking and carrying  \\ \hline
					2021 & \textbf{BUAA-Duke-Gait} (Ours) & 1,404 & Multiple$^b$ & Outdoor, Dynamic & Yes & No & RGB$^c$, Sil., GEI & Free walking and carrying.\\ \hline
				\end{tabular}	
				
				$^a$ All other datasets were collected in the daytime, while only the CASIA-C dataset contains images captured at night.
				
				$^b$ Due to the arbitrary trajectory of natural walking, the actual view angle between the camera and the target's walking direction can vary.
				
				$^c$ Our RGB images can be directly compared to person re-identification tasks.
				
		}}
		\label{tab:datasets}	
	\end{table*}
	
	\subsection{Appearance-based Gait Recognition}
	Appearance-based gait recognition methods usually encode pedestrian images with a deep convolutional neural network and then identify pedestrians based on the learned gait embeddings. Depending on the input data, these methods can be further divided into three categories: template-based approaches, sequence-based approaches, and set-based approaches.
	
	\subsubsection{Template-based Approaches}
	Template-based methods use CNNs to extract features from a single gait image, such as the Gait Energy Images (GEI)~\cite{Han2006} or other GEI-like template images~\cite{Aggarwal2018}. Wu et al.~\cite{Wu2017} proposed CNNs with different architectures to improve the performance of cross-view gait recognition\footnote{In a cross-view gait recognition setting, the gait sequences registered in the gallery set should be in the same view. Finally, all the cross-view combinations are considered, and the average performance is reported.}. Related studies can be found in~\cite{Shiraga2016, Zhang2016, Zhang2017}. Meanwhile, some generative models have also been proposed to transform gait images from one view to another, e.g. based on auto-encoders~\cite{Yu2017} and generative adversarial networks (GAN)~\cite{Yu,He2018,Wang2019a}. 
	
	\subsubsection{Sequence-based Approaches}
	
	Template-based representation abstracts the body motion by simple operations, for example, the average GEI. Since many motion details are lost due to this abstraction, sequence-based or video-based approaches are developed for a comprehensive investigation of motion. Specifically, sequence-based approaches make use of the silhouette of every frame instead of a single template image. The commonly used models for temporal feature extraction of gait include Long Short Term Memory (LSTM)~\cite{Feng2017, Tong2018, Zhang2019d}, and three-dimensional convolutional neural network (3D-CNN)~\cite{Xing2018, Lin2020}. Chen et al.~\cite{Chen2017} developed a CNN to aggregate gait silhouette sequence. Feng et al.~\cite{Feng2017} applied LSTM for cross-view gait recognition. Xing et al.~\cite{Xing2018} proposed a 3D-CNN for view-invariant gait recognition, in which spatial and temporal information are captured simultaneously. Li et al.~\cite{Li2019} proposed a comprehensive model with both LSTM and residual attention components for cross-view gait recognition. Since these studies make use of both frame-level spatial and temporal information, better representations of gait features are achieved.
	
	\begin{figure*}[t]
		\centering
		\includegraphics[width=\linewidth]{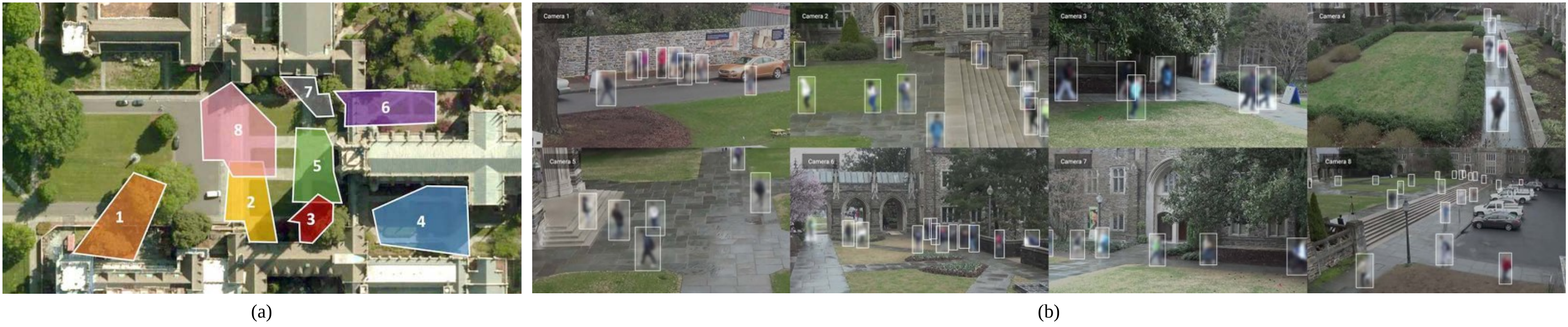}
		\caption{Eight cameras deployed on the campus of Duke University. (a) Layout of the cameras (from \cite{gou2017dukemtmc4reid}); (b) Image frames and pedestrian annotations (from \cite{ristani2016MTMC}).}
		\label{fig:dukemtmc}
	\end{figure*}
	
	\subsubsection{Set-based Approaches}
	
	Set-based approaches have been shown to achieve state-of-the-art performance on large gait datasets. Chao et al.~\cite{Chao11} regard gait as a set of discrete silhouettes rather than a continuous sequence. They believe that the silhouette appearance is a replacement for temporal information since it contains spatial information. Their GaitSet model applies a CNN to extract frame-level gait features from the silhouette and then uses a pooling operation to aggregate them into a set-level feature. Similar modules can be found in the Feature Map Pooling method~\cite{Chen2017}, the LSTM attention model~\cite{Zhang2019b}, and the Micro-motion Capture model~\cite{Fan2020}. Zhang et al.~\cite{Zhang2019b} combined CNNs for frame feature extraction with LSTM attention models. Fan et al.~\cite{Fan2020} proposed a novel parts-based model with a micro-motion capture module. 
	
	\subsection{Model-based Gait Recognition}
	Model-based methods extract gait features from human body structure~\cite{An.2020, Li.2020, Liao.2020, Li2020}. Usually, a three-dimensional body skeleton is constructed with movement patterns. Compared with appearance-based methods, model-based methods can solve the cross-view problem by simply rotating the three-dimensional body model and are considered more robust to shape changes. However, the performance of such models is limited due to the inevitable errors in body structure reconstruction. 
	
	\begin{table*}[t]
		{
			\setlength{\abovecaptionskip}{1pt}
			\renewcommand{\arraystretch}{1.3}
			\setlength{\tabcolsep}{10mm}
			\caption{Statistics of BUAA-Duke-Gait dataset.}
			{
				\setlength{\tabcolsep}{4mm}
				\label{table:stat1}
				\begin{tabular}{l|cccccccc|c}
					\hline
					\textbf{Camera \#}& \textbf{1} & \textbf{2} & \textbf{3} & \textbf{4} & \textbf{5} & \textbf{6} & \textbf{7} & \textbf{8} & \textbf{Total}\\
					\hline
					\textbf{\#Subjects}  & 817 & 761 & 391 & 320 & 434 & 717 & 456 & 528 & \textbf{1,404$^*$}\\ 
					\textbf{\#Videos} & 828  & 794  & 399 & 322  & 470 & 762 & 479 & 558  & \textbf{4,612}\\ 
					\textbf{\#Frames} & 620,308 & 653,609 & 227,653 & 296,482 & 340,253 & 826,566 & 307,140 & 351,477 & \textbf{3,623,488} \\ 
					\hline
				\end{tabular}	
			}	
			
			$^*$The total number of unique subjects.
		}
		
	\end{table*}
	
	\subsection{Video-based Person Re-Identification }
	Person re-identification (Re-ID) is a retrieval problem that searches for a particular person in non-overlapping camera views. Full color still image or video of a pedestrian can be used in person Re-ID rather than simple silhouettes as in gait recognition. Therefore, person Re-ID relies heavily on appearance information, resulting in higher recognition performance than gait recognition.
	
	Similar to gait recognition, CNN models are also widely used in person Re-ID. Some approaches have focused on mining of spatial-temporal relationships. Gu et al.~\cite{gu2020AP3D} proposed a network that aligns the adjacent frames at a pixel level to address the problem of spatial misalignment. Yan et al.~\cite{yan2020learning} constructed a graph-based network that models spatial-temporal dependencies for a person and aggregate features to address misalignment and occlusion problems. Hou et al.~\cite{BiCnet-TKS} proposed a network with two branches for spatial complementary modeling. Short-term visual clues and long-term temporal relations are both extracted in different branches and then merged to obtain video features for recognition. Ye et al.~\cite{ye2021deep} provided a comprehensive overview for person Re-ID and designed a powerful baseline model with non-local attention blocks and weighted regularization triplet loss.
	
	In general, video-based person re-identification focuses on extracting discriminative pedestrian appearance information (clothing color, etc.) with attention schemes in the spatial domain, aligning body parts in a temporal domain for improving robustness to occlusion, and fusion of spatio-temporal features. Although this framework can achieve good performance in pedestrian retrieval, it relies on appearance. Nevertheless, its effective application in open scenarios inspires us to extend gait recognition to real surveillance scenarios.
	
	\section{The BUAA-Duke-Gait Dataset}
	\label{sec:data}
	As mentioned earlier, existing gait datasets are simplified simulations of real-world scenarios. There is a lack of data for gait recognition under video surveillance, in which assumptions of user cooperation or pedestrian movement constraints are impractical. To address this issue, we construct a new gait dataset referred as the \emph{BUAA-Duke-Gait} dataset. 
	
	Given the advances in intelligent video surveillance, especially video-based person re-identification, large-scale datasets of pedestrian image sequences are becoming available. The \emph{Duke Multi-Target, Multi-Camera (DukeMTMC)} dataset~\cite{ristani2016MTMC} consists of eight 1080p videos of 85 minute duration. Each video corresponds to a surveillance camera deployed on the campus of Duke University (see Figure~\ref{fig:dukemtmc}). Based on \emph{DukeMTMC} dataset, Gou et al.~\cite{gou2017dukemtmc4reid} developed a person re-identification dataset and Wu et al.~\cite{wu2018cvpr_oneshot} developed a video-based person re-identification dataset, called \emph{DukeMTMC-VideoReID}. They copied pedestrian images from the original videos and split the dataset into 702 identities for training, 702 identities for testing, and 408 identities as distractors.
	
	Based on the pedestrian associations across cameras provided in DukeMTMC-VideoReID and the original videos in DukeMTMC, we construct a new gait dataset, \emph{BUAA-Duke-Gait} dataset, in which we extract the binary gait silhouette sequence for each subject from their original chromatic videos. In this way, the appearance information of pedestrians is removed, which makes it suitable for a gait recognition task. In addition, our new dataset provides not only a novel benchmark of gait recognition in the wild but also enables a consistent comparison between binary silhouette-based gait recognition and chromatic image-based person identification.
	
	In the following, we will describe the proposed \emph{BUAA-Duke-Gait} dataset in detail, including gait silhouette extraction, data statistics and the new challenges it provides. 
	
	\begin{figure*}[t]
		\centering
		\includegraphics[width=\linewidth]{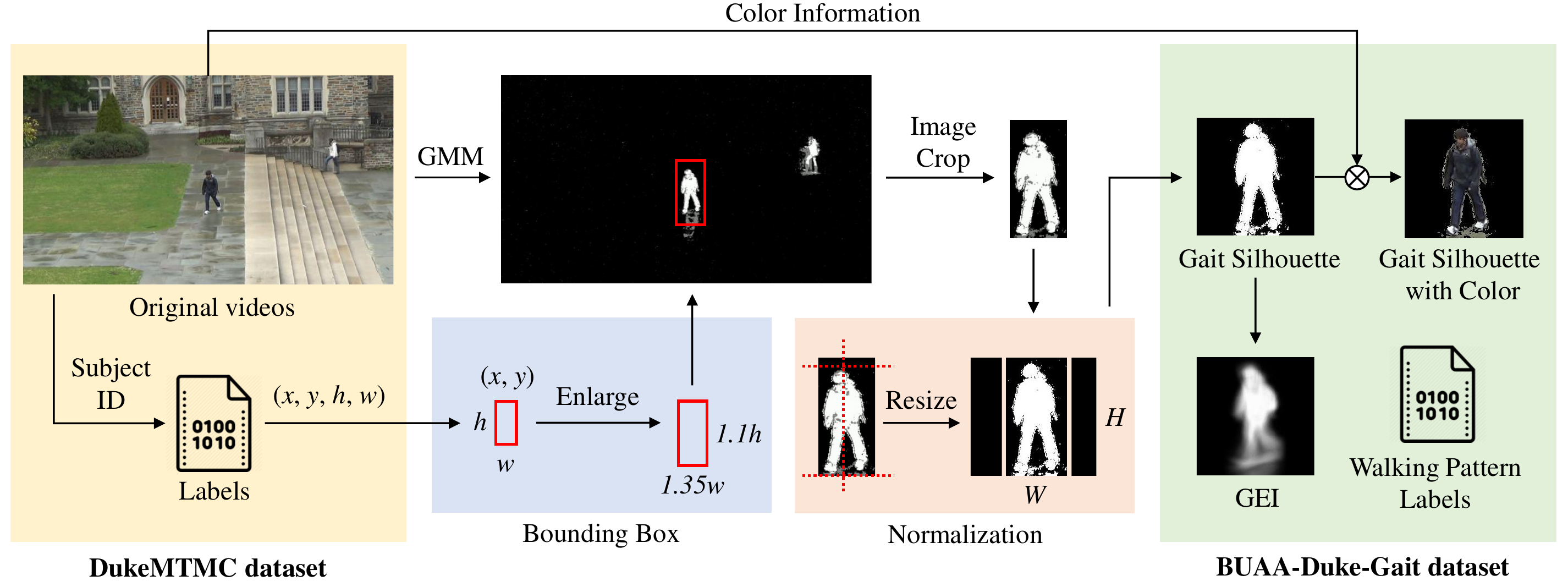}
		\caption{Construction of \textit{BUAA-Duke-Gait} dataset. Binary gait silhouettes, silhouettes with color and three different types of GEI are contained in \textit{BUAA-Duke-Gait}.}
		\label{fig:dukegait-data}
	\end{figure*}
	
	\subsection{Gait Silhouette Extraction}
	We extract gait silhouettes for each subject using the binary mask given by background subtraction and the body bounding box provided in \textit{DukeMTMC-VideoReID}. The full procedure of data acquisition is shown in Figure~\ref{fig:dukegait-data}. The key steps include background subtraction, pedestrian extraction and silhouette normalization.
	
	\subsubsection{Background Subtraction}
	
	In video surveillance, where the image resolution is often limited to the extent that face cannot be accurately recognized, the appearance of body, including color and style of clothing, is often used as a reliable cue for recognition in restricted scenarios. In other words, it is effective only when the number of people in the field of view is limited and their clothing color and style are diverse. That is the basis of person re-identification. However, although full color images contain more information than silhouettes, gait recognition's focus is on the limited yet more reliable motion pattern, which has been shown to be a useful biometric in these situations. To obtain the motion pattern, the background and pedestrian appearance need to be removed from the original video. 
	
	In this work, the Gaussian Mixture Model (GMM)~\cite{zivkovic2004improved} (implemented in OpenCV~\cite{opencv}) is used for background subtraction. Some prior works~\cite{Song.2019, Zhang2019d, Li2020} suggest using state-of-the-art segmentation methods for background subtraction, such as FCN, Mask R-CNN, or other deep learning-based models. However, we found that these methods are difficult to adopt for the following reasons.
	\begin{itemize}
		\item To generate good silhouettes from new images, the deep model has to be fine-tuned with labeled pedestrian silhouettes~\cite{Song.2019}. In a surveillance scenario, it is impractical to label the silhouettes and fine-tune the segmentation model for every camera.
		
		\item Illumination variations, background clutter and occlusion make the videos very different from those captured in controlled conditions. Consequently, directly adopting silhouettes from existing constrained gait datasets is not likely to result in a satisfactory segmentation model.
		
		\item Although GMM also suffers from inaccurate and noisy output, it does preserve motion difference between foreground object and the static background, which is consistent with the nature of gait. 	
	\end{itemize}
	
	Based on these observations, we use GMM for background subtraction.
	
	\subsubsection{Pedestrian Extraction}
	After background subtraction, the semi-automatically generated pedestrian bounding boxes labeled by DukeMTMC dataset~\cite{ristani2016MTMC} are used to crop out the image of each pedestrian. For each person, their bounding box is obtained for every fifth frame. Linear interpolation is used to calculate the bounding box in the remaining frames. We found that because the original bounding boxes are rather small, pedestrians are occluded in some frames. Therefore, we enlarge the original bounding box so that the pedestrian in the expanded bounding box is not obscured. The height and width of the bounding boxes are expanded by 1.1 times and 1.35 times, respectively.
	
	\subsubsection{Silhouette Normalization} 
	\label{section:normalization}
	Extracted silhouettes are normalized based on the method similar to~\cite{iwama2012isir, Chao11}. First, the top and bottom of the silhouette regions are obtained for each frame. Then the center of the x-axis is estimated by calculating the cumulative sum of pixels over the x-axis. Finally, we normalize the size of silhouette images to preserve the aspect ratio, where the normalized height is 224 pixels. 
	
	\subsubsection{Gait Energy Images} 
	
	For a better comparison between two gait patterns, we also generate different types of Gait Energy Images (GEI). Previous works~\cite{Takemura2018, Yu2006a} apply Normalized Auto Correlation (NAC) for gait period detection, and then calculate GEI in a gait cycle according to~\cite{Han2006}
	\begin{equation}
	G(x, y)=\frac{1}{N} \sum_{t=1}^{N} B_{t}(x, y),
	\end{equation} 
	where $N$ is the number of frames in a complete cycle or several gait cycles, $t$ is the frame number, and $B_{t}$ is a gait silhouette. It is difficult to estimate gait cycles for pedestrians with complex walking status. Therefore, all frames in a gait sequence are used to generate a GEI, named \emph{GEI-Full}.
	
\begin{figure}[t]
		\centering
		\includegraphics[width=\linewidth]{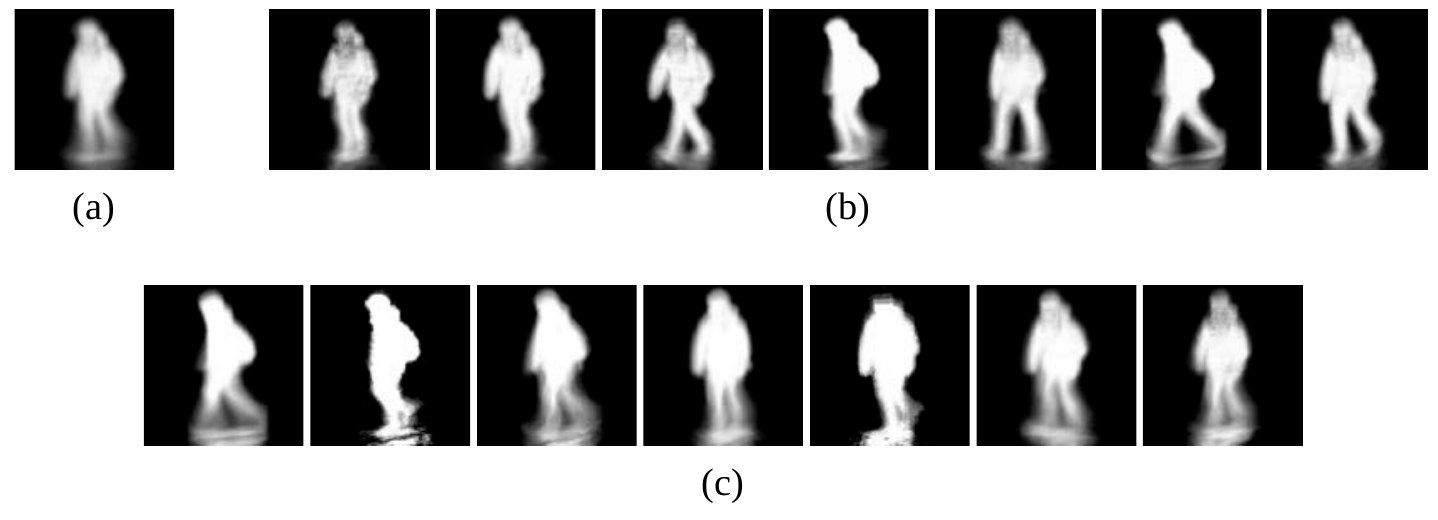}
		\caption{Examples of different types of GEI of a pedestrian walking along a curved trajectory: (a) \textit{GEI-Full}; (b) \textit{GEI-Cluster}; (c) \textit{GEI-Piecewise}. GEI-Piecewise looks better since it can keep both the leg movement information and the viewpoint information.}
		\label{fig:gei}
	\end{figure}

	Furthermore, since pedestrians walk freely in our new BUAA-Duke-Gait dataset, we also generate two other types of GEI. Inspired by~\cite{lu2013human}, K-means clustering algorithm is used before averaging all frames. For each gait sequence, all frames are grouped into seven clusters. In this way, seven GEIs can be generated for each gait sequence, named~\emph{GEI-Cluster}. To better solve the GEI generation problem, we propose a method that uses the center coordinates of a pedestrian bounding box as trajectory points. Then, the Piecewise Regression Algorithm~\cite{Kappel} is used to split this curved trajectory into multiple straight lines. Finally, we generate one GEI for each line, named \emph{GEI-Piecewise}. GEI-Piecewise can keep both the leg movement information and the viewpoint information when pedestrians are walking along curved trajectories. Exemplar images of these GEIs are shown in Figure~\ref{fig:gei}. 
	
	\begin{figure}[t]
		\centering
		\includegraphics[width=\linewidth]{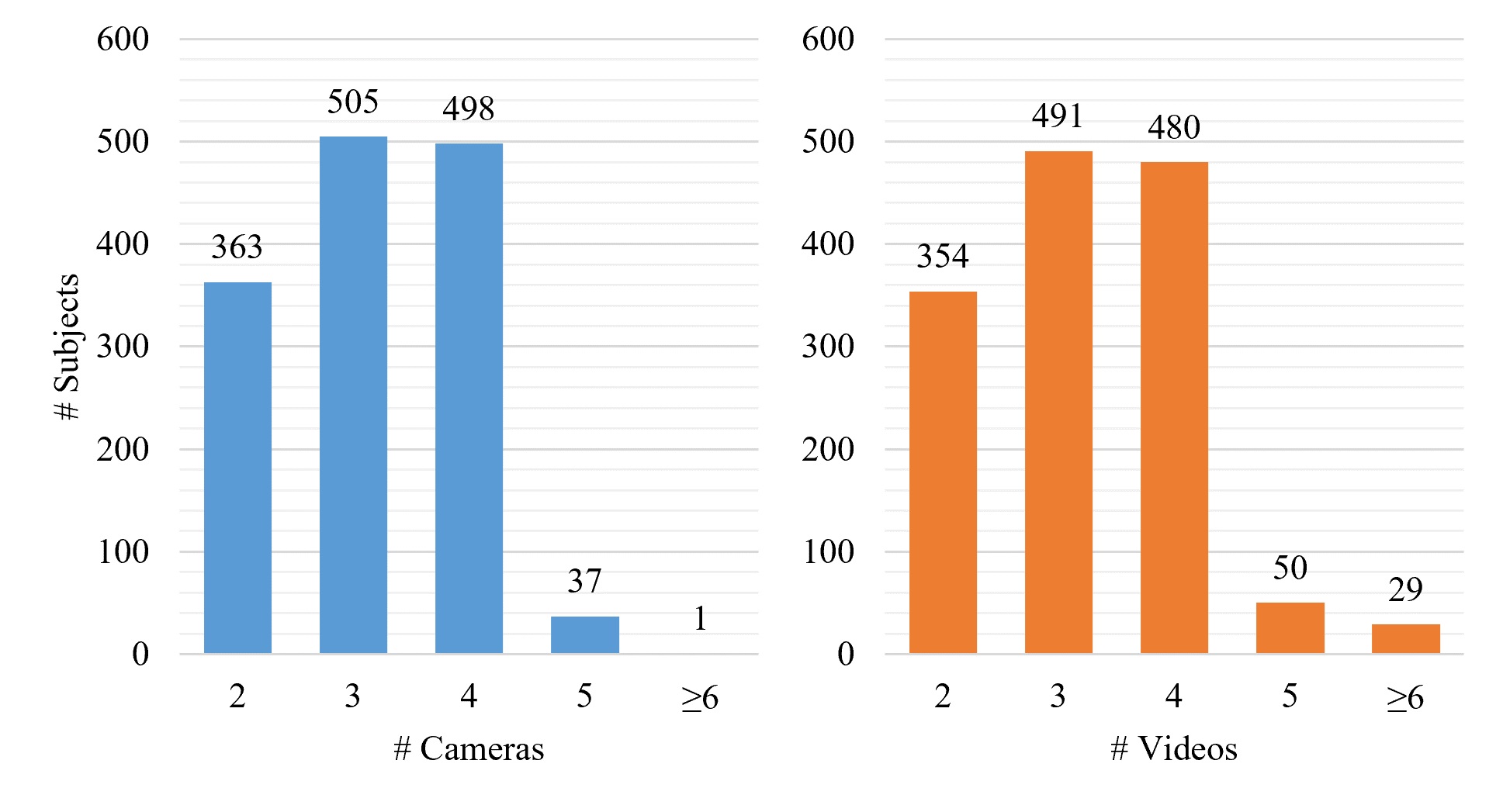}
		\caption{Statistics of the number of subjects appearing in viewpoints of different cameras and videos in \emph{BUAA-Duke-Gait} dataset.}
		\label{fig:stat-subject}
	\end{figure}
	
	\subsection{Statistics}
	The proposed gait dataset is based on DukeMTMC-VideoReID, which includes 1,404 subjects and 4,612 videos captured from eight cameras. The statistics of BUAA-Duke-Gait are shown in Table~\ref{table:stat1}. The training and test sets are strictly consistent with DukeMTMC-VideoReID. However, because the authors of DukeMTMC-VideoReID did not provide the bounding box for the distractor set, it is not included in BUAA-Duke-Gait. Later we will show empirically that the performance difference between using the distractor set or not is very small. It is worth noting that pedestrians always appear in more than one camera. Finally, over three million frames of pedestrians are obtained after our video processing. Figure~\ref{fig:stat-subject} shows the statistics of subjects appearing in different cameras and videos. Most people appear in two to four cameras and also have fewer than four videos/tracklets.
	
	\subsection{Challenges}
	\label{sec:challenge}
	Compared with the previous gait datasets captured in a controlled environment, pedestrians in the BUAA-Duke-Gait dataset walk freely with accompanying persons and a cluttered background. Therefore, it presents many new challenges to gait recognition, such as viewpoint variations, complex walking patterns, and low silhouette quality.
	
	\subsubsection{Variations in Viewpoint}
	
	Typical gait datasets assume that the walking direction is well-controlled and the camera positions are fixed. The angular difference between the camera's optical axis and the walking direction (viewpoint) is also fixed. In other words, the ``cross-view'' is limited to multiples of 15\degree. The variation in viewpoints in such datasets can be considered to be minimal. On the other hand, in the proposed BUAA-Duke-Gait dataset, the viewpoint difference can cover any possible angle irrespective of ``multi-view'' or ``cross-view''. That is a challenge of real-world gait recognition. 
	
	\subsubsection{Complex Walking Pattern}
	\label{sec:walkingstatus}
\begin{figure}[t]
		\centering
		\includegraphics[width=\linewidth]{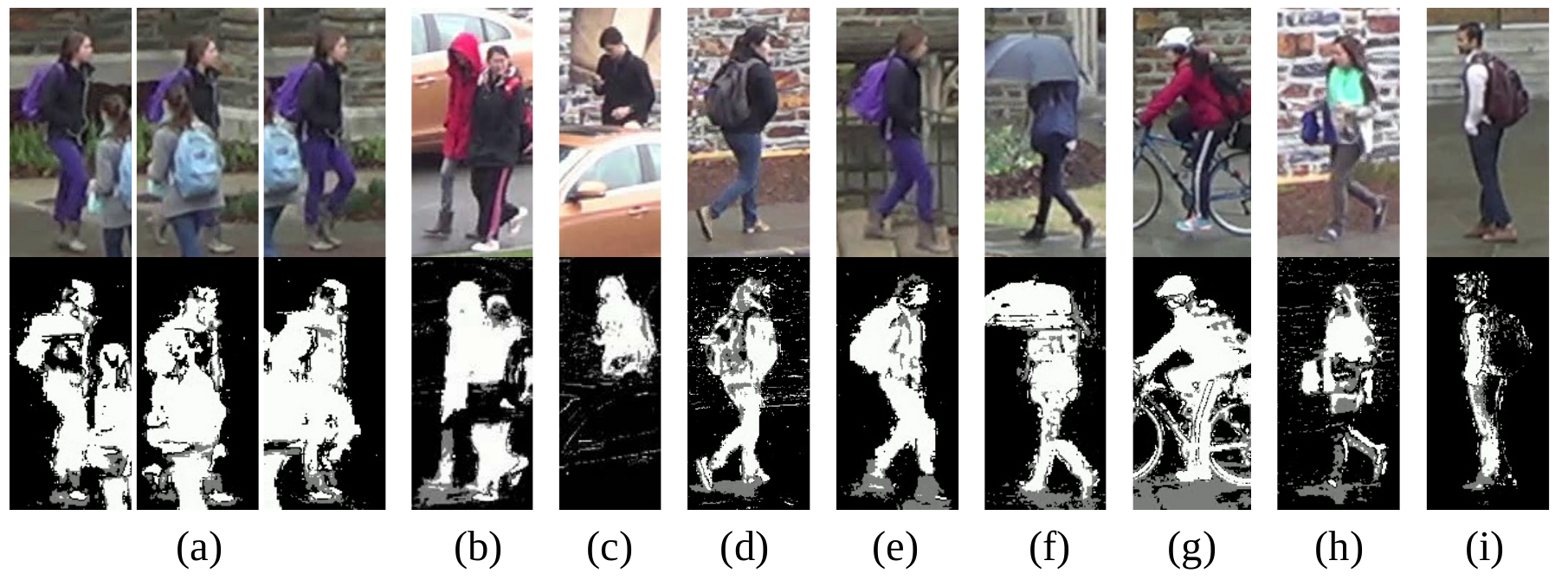}
		\caption{Examples of subjects with complex walking patterns. Top row: original images. Bottom row: extracted gait silhouettes. Walking status: (a) Crossing; (b) Companion; (c) Occlusion; (d) Bag; (e) Stairs; (f) Umbrella; (g) Bicycle; (h) Carrying an object; (i) Standing.}
		\label{fig:specialwalking}
	\end{figure}

	In our BUAA-Duke-Gait dataset, pedestrians can walk in an arbitrary pattern. Some examples are displayed in Figure~\ref{fig:specialwalking}, and the corresponding statistics are given in Table~\ref{table:walkingcondition}. Such walking patterns have not been completely investigated in previous gait datasets except for \emph{bag}~\cite{uddin2018isir}. Such complex walking patterns make the proposed dataset very challenging. It is not surprising that the performance of existing gait recognition methods deteriorate on this dataset.
	
	\begin{table}[t]
		\setlength{\abovecaptionskip}{1pt}
		\renewcommand{\arraystretch}{1.3}
		\setlength{\tabcolsep}{3mm}
		
		\caption{Statistics of the number of subjects and videos with complex walking pattern and their percentages.
		}
		\centering
		\begin{tabular}{c|cc|cc}
			\hline
			\textbf{Walking Pattern} & \multicolumn{2}{c|}{\textbf{\# Subjects}} & \multicolumn{2}{c}{\textbf{\# videos}} \\ \hline
			Crossing                    & 1379               & 98.22\%              & 3492               & 75.72\%               \\ 
			Companion                   & 345                & 24.57\%              & 1020               & 22.12\%               \\ 
			Occlusion                   & 777                & 55.34\%              & 1013               & 21.96\%               \\ 
			Bag                         & 1230               & 87.61\%              & 3848               & 83.43\%               \\ 
			Stairs                      & 723                & 51.50\%              & 750                & 16.26\%               \\ \
			Umbrella                    & 39                 & 2.78\%               & 114                & 2.47\%                \\ 
			Bicycle                     & 9                  & 0.64\%               & 25                 & 0.54\%                \\ 
			Carrying object                    & 736                & 52.42\%              & 1952               & 42.32\%               \\ 
			Standing                    & 90                 & 6.41\%               & 140                & 3.04\%   \\ \hline
		\end{tabular}
		\label{table:walkingcondition}
	\end{table}
	
	\emph{Crossing} walking pattern represents two or more pedestrians walking in opposite directions and contains a short duration where their silhouettes overlap. It is a common occurrence in real environments but has not been studied in prior art. In addition, \emph{companion} walking pattern refers to two or more people walking in the same direction in pairs or groups. In this case, the subject is obscured for a longer period compared to crossing. Meanwhile, \emph{occlusion} refers to pedestrians obscured by stationary objects, such as cars, street signs, etc. In all the three above-mentioned cases, the pedestrian images exhibit occlusion during walking, so real-world gait recognition is more challenging than controlled gait acquisition.
	
	Pedestrians carrying \emph{bags}, walking up or down \emph{the stairs}, holding \emph{umbrellas} or other objects can also interfere with the body shape and walking habit of a pedestrian. Finally, a small number of subjects may just be standing still, which results in failure of GMM for segmentation leasing to poor quality silhouettes. Even though these special walking patterns may account for only a small percentage of cases the dataset, they can lead to recognition performance degradation. Therefore, it is necessary to use examples of these real gait silhouettes to train a gait recognition system.
	
	\subsubsection{Low Silhouette Quality}
\begin{figure}[t]
		\centering
		\includegraphics[width=\linewidth]{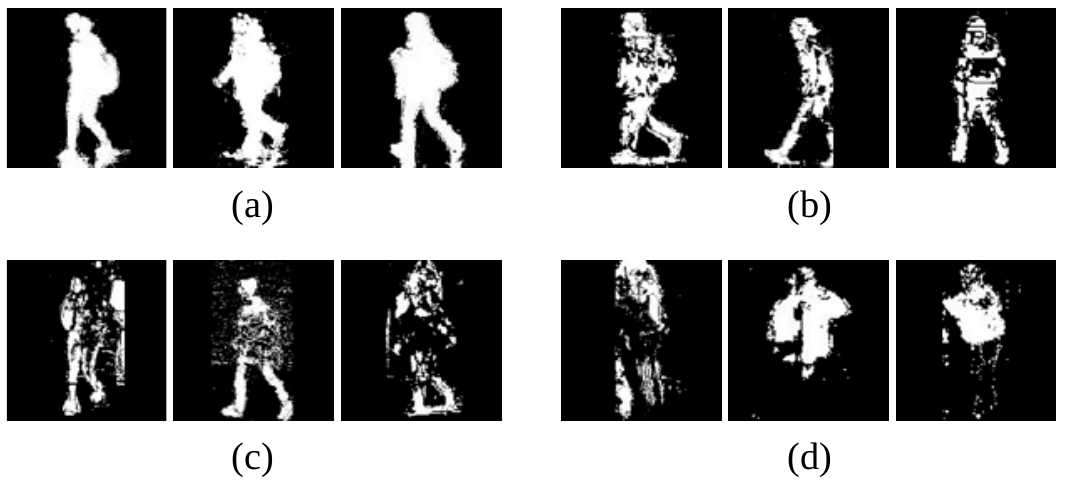}
		\caption{Gait silhouettes with different image qualities on BUAA-Duke-Gait dataset: (a) Excellent: silhouettes are clear with less image noise; (b) Good: silhouettes keep body shapes well but contain image noise; (c) Medium: a small part of body is missing; (d) Poor: more than half of body is not visible.}
		\label{fig:quality}
	\end{figure}

	Apart from the complex walking pattern, low silhouette quality caused by background clutter will also degrade the performance of gait recognition systems. Figure~\ref{fig:quality} shows the extracted gait silhouettes with different quality levels. Most of the gait silhouettes are of good quality, where the body shapes are relatively well defined. Several silhouettes are of medium quality containing some missing body parts. If more than half of the body is not visible, the silhouette is classified as poor quality. Due to weather conditions (e.g., rain and wind), camera motion, and high ISO noise caused by dim light, error in background subtraction is inevitable. It is another challenge that was not investigated in prior art.

	\begin{figure*}[t]
		\centering
		\includegraphics[width=\linewidth]{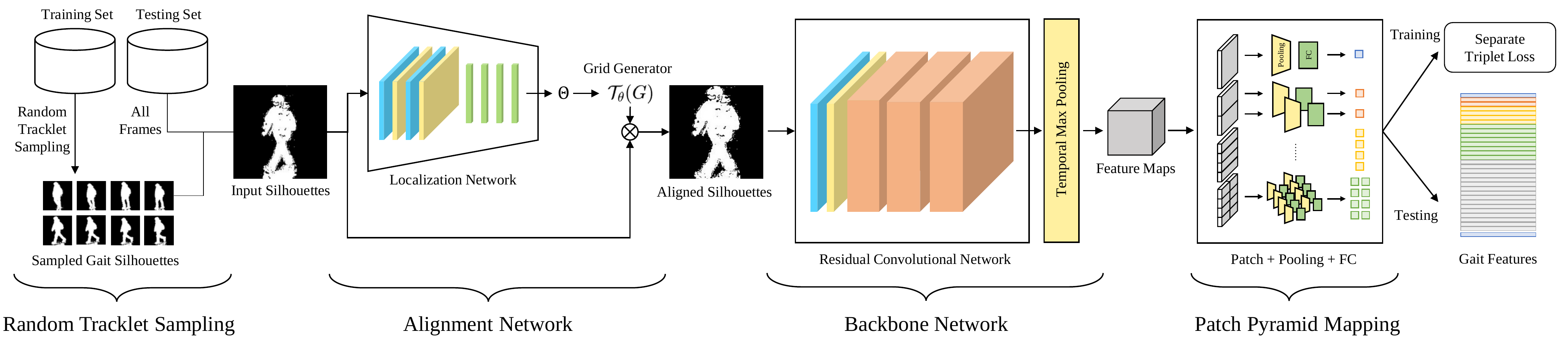}
		\caption{Framework of our proposed method, RealGait. The framework consists of four key components: random tracklet sampling, an alignment network, a backbone network, and a Patch Pyramid Mapping.}
		\label{fig:framework}
	\end{figure*}
	
	\section{Method}
	\label{sec:method}
	We have developed a novel method, named \emph{RealGait} for gait recognition for surveillance scenarios. RealGait consists of four key components: 1) random tracklet sampling in the training phase; 2) an unsupervised image alignment network; 3) a backbone network for frame-level feature extraction, and 4) a Patch Pyramid Mapping (PPM) module for local feature aggregation. The framework of the proposed method is illustrated in Figure~\ref{fig:framework}. The details are presented below.
	
	\subsection{Random Tracklet Sampling} 
	
	\begin{figure}[t]
		\centering
		\includegraphics[width=\linewidth]{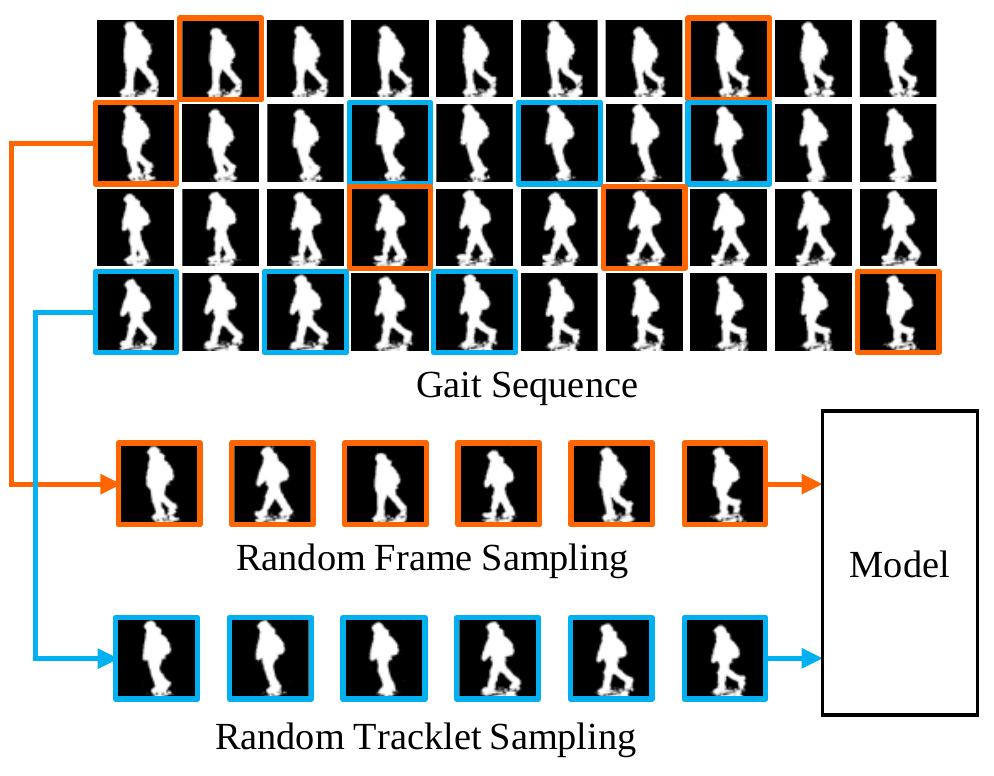}
		\caption{Illustration of random frame sampling and random tracklet sampling with $n = 40, m = 6, l = 3, s = 2$. $n$ and $m$ are the number of frames in a gait silhouette sequence and the number of random sampled frames, respectively. $l$ is the number of frames in a sampled tracklet, and $s$ is the size of the sampling steps. In random frame sampling, the period of a single instance is just a point, while in random tracklet sampling it can be an interval.}
		\label{fig:sampling}
	\end{figure}

	\label{section:sampling}
	Existing sequence-based gait recognition approaches typically use a gait sequence with fixed length as the input. Chao et al.~\cite{Chao11} noted that since body silhouettes contain the spatial information of pedestrians, which is supposed to be stable during gait cycles, the temporal continuity can be ignored. Based on this idea, in training, a gait sequence with $n$ frames is represented by $m$ random sampled frames rather than the complete tracklet, called \textit{Random Frame Sampling} (RF):
	
	\begin{equation}
	\mathcal{X}_{RF} = [x_{r1}, x_{r2}, ..., x_{rm}].
	\end{equation} 
	
	Here $r1, r2, ..., rm$ are all independent random numbers in $[1, n]$. Since this sampling strategy balances the input size and the representativeness of gait patterns, it showed improvements in conventional gait benchmarks. However, we find that in real-world scenarios, the walking patterns of a pedestrian are more complex, and the crossings and occlusions usually occur instantaneously. Meanwhile, frames with strong noise and low quality always appear in various scenarios. Even in the same sequence, individual gait cycles can be different due to the change of environment, walking patterns, and noise. Increasing the period of a single sample can better filter out such factors. For example, if a pedestrian is obscured in one frame, it is possible that the obscured area will be recovered in the subsequent frames. Therefore, these frames are spatially and temporally correlated. If a random frame sampling is used, the correlation between these frames is difficult to be learned.
	
	To this end, we propose a novel data sampling method, named \textit{Random Tracklet Sampling} (RT), as illustrated in Figure~\ref{fig:sampling}. Each time we sample several frames at equal intervals starting from a random frame in a sequence instead of just sampling random frames. In this way, we sample gait data at the tracklet-level and obtain a random frame preserving its information in a small time interval. Formally, suppose there are $n$ frames in a gait sequence $X = \{x_i| i = 1, 2, ..., n\}$. A tracklet $\mathcal{X}_r$ is a short clip of $X$ containing $l$ frames with step size $s$ where $n \geq (ls-s+1)$:
	
	\begin{equation}
	\mathcal{X}_r = [x_r, x_{r+s}, ..., x_{r+(l-1)s}]
	\end{equation}
	where $r$ is a random number in $[1, n-ls-s+2]$. Finally, $u$ tracklets are selected as the model input till the total frame number reaches $m$, satisfying $ul = m$:
	
	\begin{equation}
	\mathcal{X}_{RT} = \mathcal{X}_{r1} \cup \mathcal{X}_{r2} \cup ... \cup \mathcal{X}_{ru}
	\end{equation} 
	
	Although a tracklet contains continuous frames in RT, these frames can also be regarded as a set similar to~\cite{Chao11}. This new sampling method strikes a balance between the robustness and representativeness of gait feature expression.
	
	\begin{table}[t]
		\setlength{\abovecaptionskip}{1pt}
		\renewcommand{\arraystretch}{1.3}
		\setlength{\tabcolsep}{5mm}
		\caption{Architecture of the localization network. The ReLU activation functions are skipped in this table. The strings following each convolutional layer indicate the size of the filters and dimensions of the feature maps.}
		\centering
		\begin{tabular}{c|c}
			\hline
			\textbf{Layer}  & \textbf{Architecture} \\
			\hline
			Convolutional Layer 1 			& $7\times7$, 16, stride 2, padding 1\\	
			Max Pooling Layer 1 	& $2\times2$, stride 2\\
			Convolutional Layer 2 			& $7\times7$, 32, stride 2, padding 1\\	
			Max Pooling Layer 2 	& $2\times2$, stride 2\\
			Fully Connected Layer 1			& 7200 to 512\\
			Fully Connected Layer 2			& 512 to 128\\
			Fully Connected Layer 3			& 128 to 32\\
			Fully Connected Layer 4			& 32 to 6\\
			\hline
		\end{tabular}
		\label{table_stn}
	\end{table}
	
	\subsection{Alignment Network}
	In surveillance scenarios, the pedestrian silhouettes are difficult to align using the simple algorithm described in Section~\ref{section:normalization}, which is also the common choice in prior art. Therefore, a specific network is added for unsupervised image/frame alignment. A Spatial Transformer Network (STN)~\cite{Jaderberg2015} is proposed for image cropping and rotation. It is inserted into the CNN architecture, similar to its use in person re-identification~\cite{Zheng2019}. 
	
	We apply STN to regress a set of affine transformation parameters for alignment. The learned parameters are used to produce an image grid for image localization. At first, the original gait silhouette is input into STN for predicting an affine transformation matrix $\mathrm{A}_\theta$ 
	
	\begin{equation}
	\mathrm{A}_\theta=
	\left[
	\begin{array}{ccc}
	\theta_{11} & \theta_{12} & \theta_{13}\\
	\theta_{21} & \theta_{22} & \theta_{23}\\
	\end{array}
	\right],
	\end{equation}
	which allows cropping, translation, and isotropic scaling. Then, the point-wise transformation is defined as
	\begin{equation}
	\left(\begin{array}{c}{x_{i}^{s}} \\ {y_{i}^{s}}\end{array}\right)=\mathcal{T}_{\theta}\left(G_{i}\right)=\mathrm{A}_{\theta}\left(\begin{array}{c}{x_{i}^{t}} \\ {y_{i}^{t}} \\ {1}\end{array}\right),
	\end{equation}
	where $\mathcal{T}_{\theta}$ is a 2D affine transformation with parameter $\theta$, $(x_{i}^{t},y_{i}^{t})$ are the target coordinates of the regular grid in the output image, and $(x_{i}^{s},y_{i}^{s})$ are the source coordinates in the input image that define the sample points. Detailed architecture of this localization network is provided in Table~\ref{table_stn}.
	
\begin{figure}[t]
		\centering
		\includegraphics[width=\linewidth]{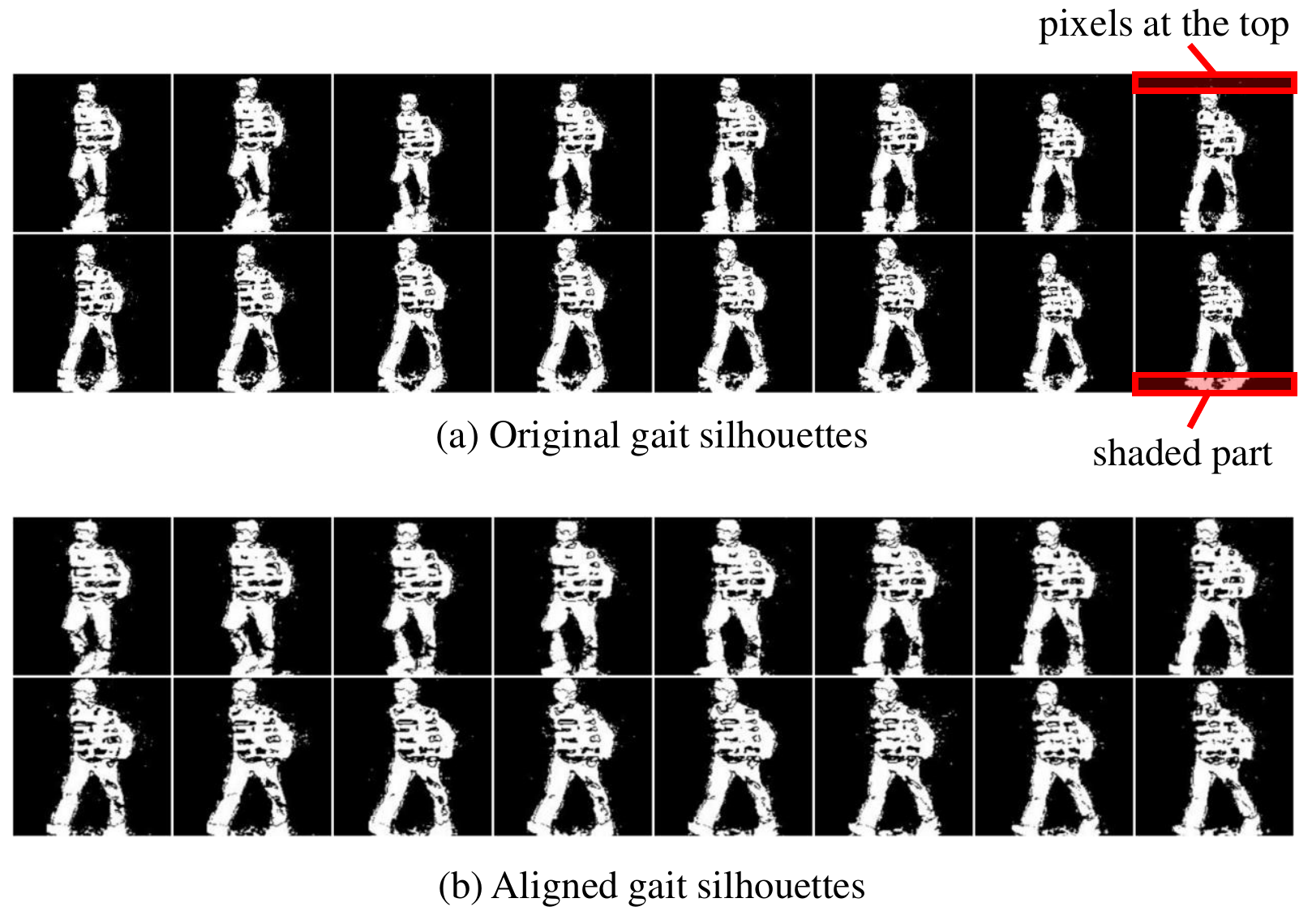}
		\caption{Exemplar results of the alignment network. Black pixels at the top and shadow pixels at the bottom of the frame are both cropped out by this alignment network.}
		\label{fig:stn}
	\end{figure}

	The visualization of this alignment network is shown in Figure~\ref{fig:stn}. The pedestrian silhouettes are not well aligned due to noise in the pre-processing step, which is corrected by the alignment network. The black pixels at the top and the shaded part at the bottom of the silhouettes are both cropped out during alignment.
	
	\subsection{Backbone Network}
	In prior art on gait recognition, shallow CNN architectures are used as the backbone for feature extraction since a binary gait silhouette is quite simple compared with the chromatic image used in other computer vision tasks. However, in our work, the gait silhouettes are more complex due to complex walking patterns and low image quality described in Section~\ref{sec:challenge}. Thus, a deeper architecture, i.e. the residual convolutional network (ResNet)~\cite{He2016}, is adopted. Table~\ref{table_resnet} shows the details of the backbone used in our framework. 
	
	After applying a residual convolutional network, a temporal max-pooling layer is applied to compress the several feature maps into one as in~\cite{Chao11}. The basis of this idea is that the gait cycle is periodic. However, it is not necessarily true when dealing with the gait pattern of random walk. However, in this paper, we keep this temporal pooling operation because of the proposed random tracklet sampling methods introduced in Section~\ref{section:sampling}.
	
	\begin{table}[t]
		\setlength{\abovecaptionskip}{1pt}
		\renewcommand{\arraystretch}{1.4}
		\setlength{\tabcolsep}{10mm}
		\caption{Architecture of our backbone network, where the layers are similar to the ResNet-18 without the classification layer and the last basic block. The activation functions ReLU and the batch norm layers are skipped. The brackets show building blocks that are the same as the architecture of ResNet-18. The strings following each convolutional layer represent the size of the filters followed by the dimensions of the feature maps.}
		\vspace{2mm}
		\setlength{\tabcolsep}{10mm}
		\centering
		\begin{tabular}{c|c}
			\hline
			\textbf{Layer} & \textbf{Architecture} \\
			\hline
			Convolutional Layer	& $7\times7$, 64, stride 2\\
			Max Pooling Layer  & $3\times3$, stride 2\\	
			
			Basic Residual Block 1	& $\begin{bmatrix} 3\times3, 64 \\ 3\times3, 64 \end{bmatrix}\times2$\\
			
			Basic Residual Block 2	& $\begin{bmatrix} 3\times3, 128 \\ 3\times3, 128 \end{bmatrix}\times2$\\
			
			Basic Residual Block 3 	& $\begin{bmatrix} 3\times3, 256 \\ 3\times3, 256 \end{bmatrix}\times2$\\	
			\hline
		\end{tabular}
		\label{table_resnet}
	\end{table}

	\subsection{Patch Pyramid Mapping}
\begin{figure}[t]
		\centering
		\includegraphics[width=\linewidth]{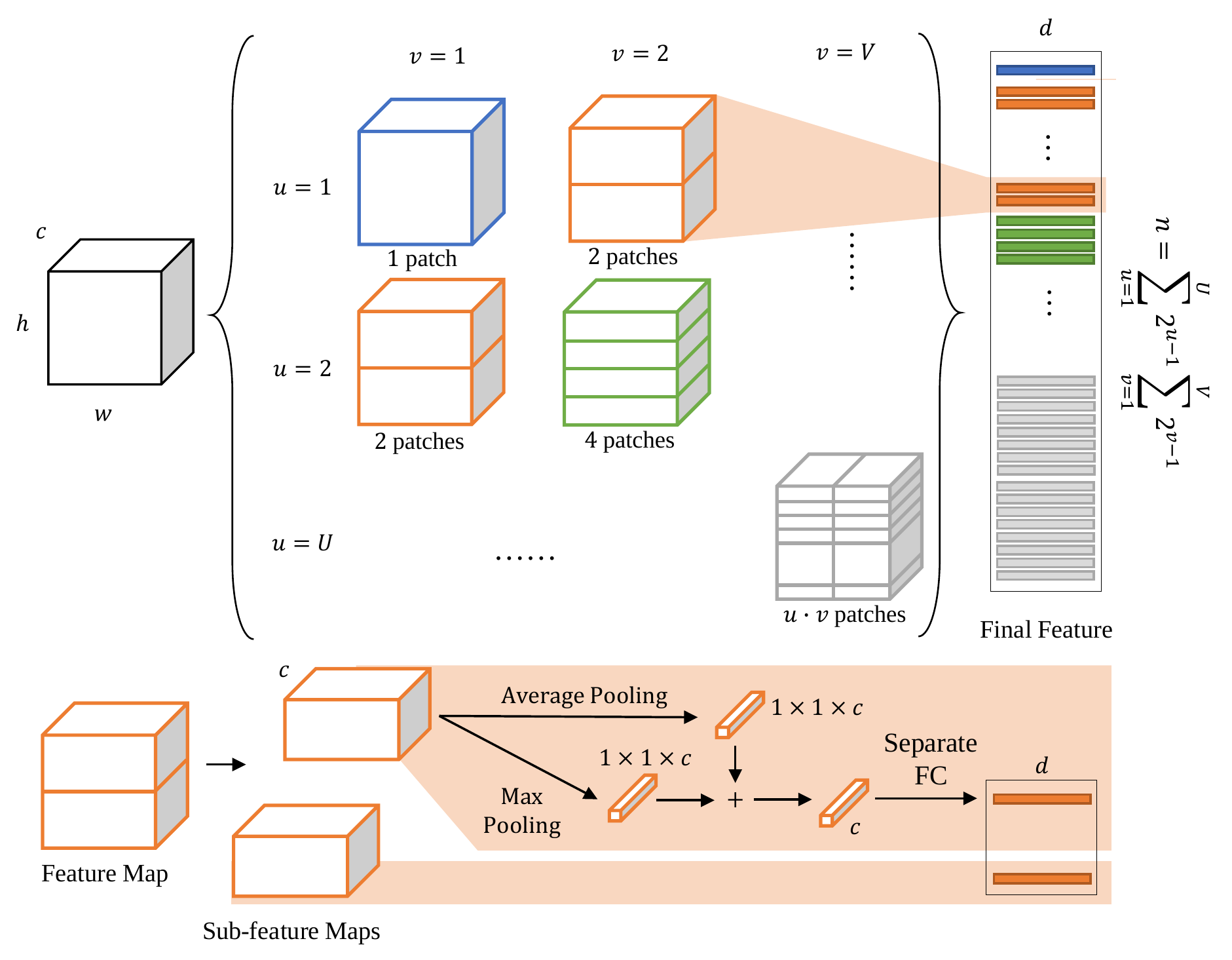}
		\caption{Structure of Patch Pyramid Mapping.}
		\label{fig:gpm}
	\end{figure}

	Previous works have shown that the part-based modules can considerably improve the performance of gait recognition, {\textit{e}.\textit{g}.} the Horizontal Pyramid Mapping~\cite{Song.2019}, Horizontal Pooling~\cite{Fan2020}, and Local Feature Extraction Module~\cite{Zhang2019b}. Following these works, we extend Horizontal Pyramid Mapping (HPM)~\cite{Song.2019} to Patch Pyramid Mapping (PPM), where not only horizontal local features but also vertical features are extracted (see Figure~\ref{fig:gpm}).
	
	PPM splits input feature maps of size $(h \times w \times c)$ into several sub-feature maps along both height and width dimensions. With scale $u \in \{1, 2,...,U\}$ and $v \in \{1, 2,...,V\}$, there are totally $U \times V$ division options. In each division option, the feature map is split into $W = 2^{u-1} \times 2^{v-1}$ sub-feature maps. When $W <= h$, we only split feature maps along height dimension, similar to HPM. The width dimension is also considered for the remaining division options where $W > h$. In this situation, the feature map is firstly split into $h$ strips along height dimension, and then $W/h$ strips along width dimension. This division method is designed because we believe that horizontal local features are more important than vertical local features. In this work, we set $U=V=4$.
	
	After patch division, a Global Average Pooling (GAP) layer and a Global Max Pooling (GMP) layer are used to downscale the sub-feature maps into $1 \times 1 \times c$. These sub-feature maps of size $c$ are mapped into a new $d$-dimensional feature. Finally, all sub-feature maps are concatenated as the final feature for similarity comparison in the test phase.

	\section{Experiments}
	\label{sec:exp}
	
	\subsection{Dataset}
	\label{section:dataset}
	
	The details of our BUAA-Duke-Gait dataset have been provided in Section~\ref{sec:data}. We follow the protocols of the DukeMTMC-VideoReID dataset in~\cite{Zheng_2017_ICCV}. There are a total of 1,404 subjects and half of them are used as the training set and the remaining in the test set. We select one video for each subject in the test set as a probe while others are in the gallery. There are 2,319 videos with 702 identities in the gallery set. All input images are resized to $256 \times 256$ to adapt our model.
	
	To further validate the proposed method, we also conduct experiments on GREW dataset~\cite{zhu2021gait}. This dataset consists of 26,345 subjects from 882 cameras in open environments with silhouettes size $64 \times 64$. We follow the evaluation protocol provided in~\cite{zhu2021gait}. There are 20,000 identities for training and 6,000 for testing. Each subject in the test set has 4 sequences, 2 for probe set and 2 for gallery set. The distractor set is removed in this work since we start from a simple experimental setting.

	\subsection{Implementation Details}
	For our proposed model, we initialize the weights of the backbone network according to the \emph{He initialization} ~\cite{he2015delving}, and others with \emph{Xavier initialization}~\cite{glorot2010understanding}. The Adam algorithm~\cite{KingmaD2014} is implemented for optimization. In training phases, batch all triplet loss~\cite{HermansBeyer2017Arxiv} with margins equal to 0.2 is used for each sub-features after PPM, similar to~\cite{Chao11}. In the BUAA-Duke-Gait dataset, since most subjects have less than four videos, we train the model with $p = 16, k = 2$ in a batch size of $p \times k$. The learning rate is $1e^{-4}$ in the first 150,000 iterations and then is changed into $1e^{-5}$ for the remaining 100,000 iterations. 
	
	For experiments on GREW dataset, since silhouettes with size $64 \times 64$ are pre-processed and provided, we slightly change our model to adapt this dataset. The Alignment Network module is removed because silhouettes of pedestrians are aligned automatically according to~\cite{zhu2021gait}. In addition, we change the size of stride in the Basic Residual Block 2 and Block 3 to $stride = 1$. The learning rate is $1e^{-4}$ in the first 250,000 iterations and $1e^{-5}$ for the remaining 350,000 iterations. 

	\subsection{Evaluation Metrics}

	Average rank-$n$ accuracy is utilized to evaluate the recognition performance of a method. Rank-$n$ accuracy is defined as the percentage of identity predictions where the top-$n$ prediction matches the ground-truth identity label. If an experiment contains more than one scene, we average the rank-$n$ accuracy of each scene to calculate the average rank-$n$ accuracy as the final evaluation metric. For most of the experiments, average rank-1 accuracy is applied since it is the most popular metric in gait recognition, while rank-5, rank-10, and rank-20 are also calculated for better comparison on GREW dataset.
	
	In the experimental settings where a pedestrian appearing in the probe camera may not appear in the gallery camera, we consider this recognition task as an open-set identification problem. Rank-1 Detection and Identification Rate (DIR) at a fixed False Acceptance Rate (FAR) is reported for this open-set identification problem. When FAR equals 100\%, DIR becomes a rank-1 accuracy in cross-view closed-set settings, which is usually reported in previous gait recognition works.

	\subsection{Comparison to Gait Recognition Approaches}
	\subsubsection{Settings} 
	\label{sec:gait_setting}
	
	As defined in previous work~\cite{Wu2017}, in a cross-view gait recognition task, the gait sequences registered in the gallery set should be from a specific view, while in a multi-view recognition task, sequences of all views are registered in the gallery set. In BUAA-Duke-Gait, due to free walking, it is hard to estimate walking view angles for each pedestrian. Therefore, we expanded these two definitions to scene-level. Following the settings of conventional gait recognition, we conduct two identification tasks: multi-scene recognition and cross-scene recognition, respectively.
	
	In a multi-scene recognition task, for each identity, we select one video as a probe according to protocols introduced in Section~\ref{section:dataset}, while all other videos with different cameras or different identities are registered as gallery videos. In a cross-scene recognition task, all different pairs of eight cameras are selected as probe and gallery. In each scene pair, videos in one scene are registered as gallery, and videos from the other scene are used as probe. In this series of experiments on BUAA-Duke-Gait dataset, the default setting is multi-scene recognition except for specific notation of cross-scene.

	\subsubsection{Main Results}
	We first compare our method with state-of-the-art gait recognition methods on BUAA-Duke-Gait dataset with both multi-scene and cross-scene tasks in Table~\ref{table_comparsion_Duke-Gait} and GREW in Table~\ref{table_comparsion_GREW}.
	
	\textbf{Results on BUAA-Duke-Gait dataset.}
	Since the BUAA-Duke-Gait dataset is a new dataset, we could compare it to only the three recent state-of-the-art methods which provided their codes. From Table~\ref{table_comparsion_Duke-Gait}, both results show that, compared with their results on conventional gait datasets, i.e. CASIA-B~\cite{Yu2006a} and OU-MVLP~\cite{Takemura2018}, the performance of three state-of-the-art methods drops dramatically on BUAA-Duke-Gait (from over 95\% on CASIA-B and over 87\% on OU-MVLP to less than 62\% on BUAA-Duke-Gait). These results show that the gait recognition models designed for the controlled scenarios cannot perform well on real-world data like BUAA-Duke-Gait dataset. The aforementioned new challenge described in Section~\ref{sec:challenge} can greatly influence the recognition performance. It is clear that the proposed method, RealGait outperforms state-of-the-arts. RealGait achieves 78.24\% rank-1 correct recognition rate in the multi-scene recognition task, compared to the best accuracy of 61.92\% for GaitGL among the three state-of-the-art methods in a multi-scene scenario, which implies that the proposed framework is effective. Similar results can be found in the cross-scene recognition task.
	
	\begin{table}[t]
		{
			\setlength{\abovecaptionskip}{1pt}
			\renewcommand{\arraystretch}{1.3}
			\caption{Average rank-1 accuracies (\%) on multi-scene and cross-scene recognition on BUAA-Duke-Gait dataset. }
			\centering	
			{
				\setlength{\tabcolsep}{6.8mm}
				
				\begin{tabular}{c|c|c}
					\hline
					\textbf{Method}  &  \textbf{Multi-scene}    & \textbf{Cross-scene$^*$} \\
					\hline
					GaitSet \cite{Chao11} &  46.72   &   42.44     \\
					GaitPart \cite{Fan2020} &  33.89   &   33.78     \\
					GaitGL \cite{lin2020learning} &   61.92  &  58.66   \\ \hline
					\textbf{RealGait} &  \textbf{78.24}   &  \textbf{70.84}  \\ \hline
				\end{tabular}
				\label{table_comparsion_Duke-Gait}
				
				$^*$In this cross-scene gait recognition task, a probe video of a subject is removed if the subject does not appears in the gallery.
		}}
	\end{table}
	
	\begin{table}[t]
		\setlength{\abovecaptionskip}{1pt}
		\renewcommand{\arraystretch}{1.3}
		\caption{Average rank-$n$ accuracies (\%) on GREW dataset. Results of GaitSet and GaitPart are from~\cite{zhu2021gait}. }
		\centering	
		\setlength{\tabcolsep}{3.2mm}
		
		\begin{tabular}{c|c|c|c|c}
			\hline
			\textbf{Method}  &  \textbf{Rank-1 }    & \textbf{Rank-5 } & \textbf{Rank-10} & \textbf{Rank-20}\\
			\hline
			GaitSet \cite{Chao11} & 46.28 & 63.58 & 70.26 & 76.82 \\
			GaitPart \cite{Fan2020} & 44.01 & 60.68 & 67.25 & 73.47 \\
			GaitGL \cite{lin2020learning} & 47.28 & 63.56 & 69.32 & 74.18 \\
			\hline
			\textbf{RealGait} & \textbf{54.12} & \textbf{71.47} & \textbf{77.57} & \textbf{81.71} \\ \hline
		\end{tabular}
		\label{table_comparsion_GREW}
		
	\end{table}

\begin{figure}[t]
		\centering
		\includegraphics[width=0.7\linewidth]{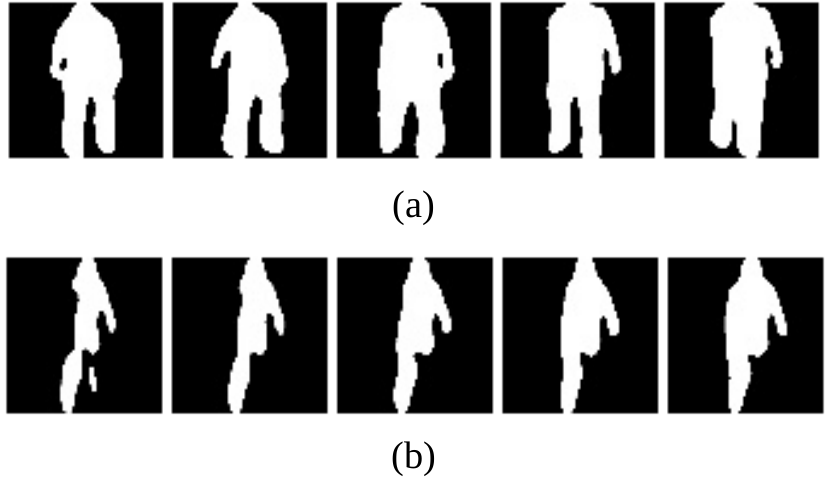}
		\caption{Silhouettes with poor quality on GREW dataset. (a) Inaccurate contours: leg widths are inconsistent; (b) Body part missing: leg and arm are completely disappeared.}
		\label{fig:quality_grew}
	\end{figure}

	\textbf{Results on GREW dataset.}
	We compare our methods with the state-of-the-art methods since these methods are baseline methods and achieved top-2 performance on GREW dataset~\cite{zhu2021gait}. From Table~\ref{table_comparsion_GREW}, our method outperforms the previous methods. This result shows that our method works well on the other gait datasets with real-world data. We also notice that silhouettes on GREW dataset are produced by an automatic segmentation method, which is different from our GMM method. This difference results in a huge distinction of image morphology on the two datasets. For example, high ISO noise and error in background subtraction can be observed on BUAA-Duke-Gait dataset (see Figure \ref{fig:quality}), while inaccurate contours and body parts missing can be observed on GREW dataset (see Figure \ref{fig:quality_grew}). Therefore, our model does not achieve high performance on GREW dataset as it on BUAA-Duke-Gait dataset.
	
	\subsubsection{Results on Open-set Cross-scene Settings}
	
	Most existing gait recognition and person Re-ID only focus on a close-set problem, i.e. probe subjects and gallery subjects are the same, which makes them unsuitable for real-world applications. We consider an open-set problem on our BUAA-Duke-Gait dataset based on cross-scene recognition settings. If a subject in a probe video does not appear in the gallery set, we regard it as an imposter, and our recognition system needs to identify imposters. The difference between this experiment and the cross-scene experiment in Table~\ref{table_comparsion_Duke-Gait} is that all subjects in the probe set are used during testing in the open-set experiment.
	
	Table~\ref{table_crossdetails} shows the results of our RealGait in an open-set cross-scene scenario on BUAA-Duke-Gait dataset. We find that the recognition performance is low in the experiments associated with the Camera \#4. From Figure~\ref{fig:dukemtmc}, the road is oriented vertically in the Camera \#4 images. This means that all pedestrians are either facing the camera or facing away from the camera. This situation is similar to the camera view angle being either 0 or 180 degrees in the previous gait datasets. Since the leg motion is unclear in the images of the front-facing pedestrians, the gait recognition performance is poor. This result and conclusion are consistent with the previous findings~\cite{Wu2017, Zhang2019d}.
	
	\begin{table}[t]
		\setlength{\abovecaptionskip}{1pt}
		\renewcommand{\arraystretch}{1.3}
		\caption{Average rank-1 DIR (\%) of RealGait in open-set cross-scene scenario on BUAA-Duke-Gait dataset.}
		\centering	
		\setlength{\tabcolsep}{4.6mm}
		
		\begin{tabular}{l|c|c|c|c|c}
			\hline
			\multicolumn{2}{c|}{} & \multicolumn{4}{c}{\textbf{Gallery All Cameras (\#1 -- \#8)}} \\ \hline
			\multicolumn{2}{c|}{\textbf{FAR (\%)}}  & \textbf{1} & \textbf{10} & \textbf{50} & \textbf{100}              \\ \hline
			\multicolumn{1}{l|}{\multirow{8}{*}{\textbf{\rotatebox{90}{Probe Camera}}}} & \textbf{\#1}     & 42.14      & 62.31       & 74.65       & 76.81                     \\ \cline{2-6} 
			\multicolumn{1}{l|}{}                                       & \textbf{\#2}     & 48.02      & 58.72       & 74.20       & 76.71                     \\ \cline{2-6} 
			\multicolumn{1}{l|}{}                                       & \textbf{\#3}     & 41.78      & 62.53       & 74.42       & 79.41                     \\ \cline{2-6} 
			\multicolumn{1}{l|}{}                                       & \textbf{\#4}     & 7.88       & 18.96       & 37.01       & 42.92                     \\ \cline{2-6} 
			\multicolumn{1}{l|}{}                                       & \textbf{\#5}     & 36.90      & 57.69       & 73.47       & 76.21                     \\ \cline{2-6} 
			\multicolumn{1}{l|}{}                                       & \textbf{\#6}     & 34.76      & 49.71       & 65.47       & 67.59                     \\ \cline{2-6} 
			\multicolumn{1}{l|}{}                                       & \textbf{\#7}     & 30.47      & 53.76       & 74.02       & 78.67                     \\ \cline{2-6} 
			\multicolumn{1}{l|}{}                                       & \textbf{\#8}     & 32.32      & 45.02       & 66.61       & 68.38                     \\ \hline
			\multicolumn{2}{c|}{\textbf{Mean}}                                             & \textbf{34.28}      & \textbf{51.09}       & \textbf{67.48}       & \textbf{70.84} \\ \hline
		\end{tabular}
		\label{table_crossdetails}
		
	\end{table}

	\subsubsection{Impact of Input Data}
	
	Since GEI is also an important and widely used gait feature in previous works, we also conduct experiments with GEI on BUAA-Duke-Gait dataset. Table~\ref{table_gei} shows the recognition performance when we replace the input data from the gait silhouette with three different GEIs and remove the temporal max-pooling layer. The resulting accuracies show that compared to averaging the gait silhouettes directly, extracting the image features first and then pooling the features provide more effective gait features. This conclusion is also consistent with the previous findings~\cite{Chao11}. Although the use of GEI can reduce the computational effort and achieve good recognition performance on small-scale datasets, the performance is poor on large-scale and unconstrained datasets.
	
	\begin{table}[t]
		\setlength{\abovecaptionskip}{1pt}
		\renewcommand{\arraystretch}{1.3}
		\caption{Average Rank-1 accuracies with GEI of multi-scene recognition on BUAA-Duke-Gait dataset.}
		\centering	
		\setlength{\tabcolsep}{6mm}
		
		\begin{tabular}{c|c|c}
			\hline
			\textbf{Method}  &  \textbf{Input}    & \textbf{Rank-1 Accuracy} \\ \hline
			\multirow{3}{*}{\textbf{{RealGait}}} & GEI-Full & 12.83\%    \\
			& GEI-Cluster & 14.28\%    \\
			& GEI-Piecewise & 16.24\%    \\ \hline
		\end{tabular}
		\label{table_gei}
		
	\end{table}

	\subsubsection{Impact of Model Architectures}
	Results of ablation experiments for model architectures are listed in Table~\ref{table_architectures_Duke-Gait}. We consider the four components of our proposed method: sampling method, alignment network, backbone network, and pyramid mapping module. The component which shows the most improvement is the backbone network. After we change the backbone model from a shallow CNN architecture in~\cite{Chao11} to several Residual Blocks as in~\cite{He2016}, the rank-1 accuracy increases from 46.72\% to 69.18\%. This result indicates that while a model with a shallow CNN can work well on a dataset in controlled scenarios because the gait silhouettes are clear and stable, in complex real-world scenarios, the features extracted by a shallow CNN do not have sufficient discriminative power.
	
	Furthermore, the introduction of alignment network and pyramid mapping modules also contributes to performance improvement. Both of them improve rank-1 accuracy by about 2\% each. We also conduct an experiment when we give the equal priority level to the vertical local feature, named PPM-V. In PPM-V, the feature map is split into $2^{u-1}$ horizontal strips along height dimension and $2^{v-1}$ vertical strips along weight dimension. The result indicates that vertical local features are useful but less important than horizontal features.
	
	At last, our new sampling method can raise rank-1 accuracy by about 3\%, and this improvement can be observed in all settings. In summary, the ablation results show that all components of our proposed method contribute to the gait recognition framework in real unconstrained surveillance scenarios. 
	
	\begin{table}[t]
		\setlength{\abovecaptionskip}{1pt}
		\renewcommand{\arraystretch}{1.3}
		\caption{Impact of model architectures on BUAA-Duke-Gait dataset. RF: Random 28 frames. RT: Random 4 tracklets $\times$ 7 frames. Convolutional layers of backbone network and HPM are from GaitSet \cite{Chao11}}
		\centering
		\setlength{\tabcolsep}{0.6mm}
		\begin{tabular}{c|c|c|c|c}
			\hline
			\begin{tabular}[c]{@{}c@{}}\textbf{Sampling}\\ \textbf{Method}\end{tabular} &
			\begin{tabular}[c]{@{}c@{}}\textbf{Alignment}\\ \textbf{Network}\end{tabular} & \begin{tabular}[c]{@{}c@{}}\textbf{Backbone}\\ \textbf{Network}\end{tabular} & \begin{tabular}[c]{@{}c@{}}\textbf{Pyramid} \\ \textbf{Mapping}\end{tabular} & \begin{tabular}[c]{@{}c@{}}\textbf{Rank-1} \\ \textbf{Accuracy}\end{tabular} \\ \hline
			
			RF &           & 6 Convolutional Layers & HPM        & 46.72\% \\
			RF &\checkmark & 6 Convolutional Layers & HPM        & 47.84\% \\
			RT &		   & 6 Convolutional Layers & HPM        & 50.77\% \\ \hline	
			
			RF &		   & 3 Residual Blocks & HPM        & 69.18\% \\
			RF &\checkmark & 3 Residual Blocks & HPM        & 71.83\% \\
			RT &\checkmark & 3 Residual Blocks & HPM        & 76.15\% \\ 
			
			\hline
			
			RF &\checkmark & 3 Residual Blocks & PPM & 73.36\% \\
			RT &           & 3 Residual Blocks & PPM & 76.43\% \\
			RT &\checkmark & 3 Residual Blocks & PPM-V      & 76.29\%   \\ 
			\hline
			RT &\checkmark & 3 Residual Blocks & PPM & \textbf{78.24\%}    \\ 
			\hline
			
		\end{tabular}
		\label{table_architectures_Duke-Gait}
	\end{table}
	
	\subsubsection{Impact of Data Sampling}
	To further validate the contributions of the proposed data sampling method, we conduct another experiment displayed in Table~\ref{table_datasampling_Duke-Gait}. In this experiment, two data sampling methods are used: random frames and random tracklets, which have been introduced in Section~\ref{section:sampling}. We investigate several settings of $u, l, s$ with the restriction that the sample image number is the same for both random frame and random tracklet.
	
	The result shows that the performance can improve by 5\% if we change the data sampling methods to the proposed random tracklet method. It provides support that the proposed sampling methods can keep the advantages of both the previous sampling methods. The features of the neighboring frames can complement the loss of information in a single frame due to pedestrian crossing and occlusion. Therefore, the proposed random tracklet method provides strong robustness in this complex scenario.
	
	\begin{table}[t]
		\setlength{\abovecaptionskip}{1pt}
		\renewcommand{\arraystretch}{1.3}
		\caption{Impact of data sampling on BUAA-Duke-Gait dataset.}
		\setlength{\tabcolsep}{5mm}
		\centering
		\begin{tabular}{c|c|c}
			\hline
			\begin{tabular}[c]{@{}c@{}}\textbf{Sampling Method}\\ \textbf{($u$ Tracklets $\times$ $l$ Frames)}\end{tabular} &
			\begin{tabular}[c]{@{}c@{}}\textbf{Sampling}\\ \textbf{(Step size $s$)}\end{tabular} & \begin{tabular}[c]{@{}c@{}}\textbf{Rank-1}\\ \textbf{Accuracy}\end{tabular} \\ \hline
			Random 28 Frames          & -      & 73.36\%  \\ \hline                             
			4 Tracklets $\times$ 7 Frames    & 2      & 77.68\%  \\
			4 Tracklets $\times$ 7 Frames    & 4      & 78.10\%  \\
			4 Tracklets $\times$ 7 Frames    & 6      & \textbf{78.24\%} \\ 
			4 Tracklets $\times$ 7 Frames    & 8      & 76.71\%  \\
			4 Tracklets $\times$ 7 Frames    & 10     & 77.68\%  \\ \hline
			3 Tracklets $\times$ 10 Frames    & 6     & 77.41\%  \\
			5 Tracklets $\times$ 6 Frames    & 6     & 76.71\%  \\
			6 Tracklets $\times$ 5 Frames    & 6     & 76.71\%  \\
			7 Tracklets $\times$ 4 Frames    & 6     & 76.01\%  \\ \hline
			
		\end{tabular}
		\label{table_datasampling_Duke-Gait}
	\end{table}
	
	\subsubsection{Impact of Walking Patterns }
	In this subsection, we will analyze the recognition performance under different complex pedestrian walking patterns described in Section~\ref{sec:walkingstatus}, shown in Table~\ref{table:resultwalkingcondition}.
	
	Among the first three walking patterns that have a relationship with occlusion during walking, our model is capable of handling. \emph{crossing} and \emph{occlusion}. In these two patterns, although the human body is partially obscured, the obscuring time period is short. The information loss will be supplemented by the resting frames in a tracklet. A similar explanation can be given for \emph{Stairs}. However, the subject is obscured for a longer period in \emph{companion} situation. The silhouettes of two or more pedestrians overlap during the whole tracklet. This leads to difficulties in individual gait information extraction by conventional pedestrian silhouettes.
	
	As for carrying objects, since the DukeMTMC-VideoReID dataset is captured on campus, most of the pedestrians are carrying bags (backpacks) in all the camera scenes. Our results show that there is little performance degradation for a gait recognition system in multi-scene when pedestrians are carrying bags or other objects. Finally, in \emph{umbrellas}, \emph{bicycle} and \emph{standing} situations, intuitively, they may impact gait recognition system performance. However, since the number of these subjects in the BUAA-Duke-Gait dataset is quite small, the impact cannot be assessed.
	
	\begin{table}[t]
		\setlength{\abovecaptionskip}{1pt}
		\renewcommand{\arraystretch}{1.3}
		\setlength{\tabcolsep}{4.5mm}
		
		\caption{Impact of complex walking pattern on BUAA-Duke-Gait dataset.}
		\centering
		\begin{tabular}{c|c|c}
			\hline
			\textbf{Walking Pattern} & \textbf{\# Subjects} & \textbf{Rank-1 Accuracy} \\ \hline
			Crossing  & 543  & 78.64\% \\ 
			Companion & 157  & 59.24\% \\ 
			Occlusion & 158  & 80.38\% \\ 
			Bag       & 615  & 78.86\% \\ 
			Stairs    & 111  & 80.18\% \\
			Umbrella  & 18  & 61.11\% \\ 
			Bicycle   & 4  & 75.00\%  \\ 
			Carrying  & 303  & 78.22\% \\ 
			Standing  & 12  & 50.00\%  \\ \hline
		\end{tabular}
		\label{table:resultwalkingcondition}
	\end{table}

	\subsection{Comparisons to Video Person Re-identification}
	\subsubsection{Settings}
	
	Although gait recognition is relevant to general pedestrian recognition tasks like person re-identification, no study on the impact of gait on these tasks has been conducted. Our new dataset makes it possible to perform consistent comparisons between gait recognition and person re-identification.
	
	For baseline experiments of re-identification methods, we strictly follow the protocols in~\cite{Zheng_2017_ICCV} and use the image tracklets provided by DukeMTMC-VideoReID. There are a total of 702 identities for training, 702 identities for testing, and 408 identities as distractors. We have 2,196 tracklets for training and 2,636 tracklets for testing and distractors. For the experiments of the proposed RealGait method, the settings are described in Section~\ref{sec:gait_setting}.
	
	\subsubsection{Main Results}
	
	Since the distractor set is missing in our BUAA-Duke-Gait, to evaluate its impact we first examine three state-of-the-art person re-identification models on original DukeMTMC-VideoReID dataset. The performance of AWG~\cite{ye2021deep}, BiCnet-TKS~\cite{BiCnet-TKS} and AP3D~\cite{gu2020AP3D} are shown in Table~\ref{table:resultreid}. These models achieve over 95\% rank-1 accuracy with distractor set (see the second column in Table~\ref{table:resultreid}).
	
	As mentioned earlier, the proposed BUAA-Duke-Gait dataset is not fully consistent with DukeMTMC-VideoReID because the distractor set is missing. To analyze the impact of this issue, we conduct experiments on the original DukeMTMC-VideoReID dataset by comparing several state-of-the-art video person re-identification methods with and without the distractors (see the third column in Table~\ref{table:resultreid}).
	
	As can be seen, the rank-1 accuracies improve by around 0.5\% to 0.7\% after removing the distractors. Since the difference in performance with and without distractors is less than 1\%, we can conclude that the influence of distractors is negligible.
	
	\begin{table}[t]
		\setlength{\abovecaptionskip}{1pt}
		\renewcommand{\arraystretch}{1.3}
		\setlength{\tabcolsep}{3.5mm}
		
		\caption{Average multi-scene recognition rank-1 accuracies of three video person Re-ID methods on both Re-ID and gait datasets with different distractor set settings.}
		\centering
		\begin{tabular}{c|cc|c}
			\hline
			\multirow{2}{*}{\textbf{Dataset}}  & \multicolumn{2}{c|}{\textbf{DukeMTMC}} & \textbf{BUAA-Duke-Gait}  \\
			& \multicolumn{2}{c|}{\textbf{VideoReID}\cite{wu2018cvpr_oneshot}} & \textbf{(Silhouette)}  \\ \hline	
			\textbf{Distractor Set}  & Yes & No &  No  \\  \hline		     
			AWG \cite{ye2021deep}        & 95.20\%  & 95.70\% & 66.70\%      \\ 
			BiCnet-TKS \cite{BiCnet-TKS} & 96.30\% & 96.87\%  & 13.81\%      \\ 
			AP3D \cite{gu2020AP3D}      & 96.60\%  & 97.30\%  & 61.10\%      \\ \hline
		\end{tabular}
		\label{table:resultreid}
	\end{table}

	\begin{figure*}[t]
		\centering
		\includegraphics[width=\linewidth]{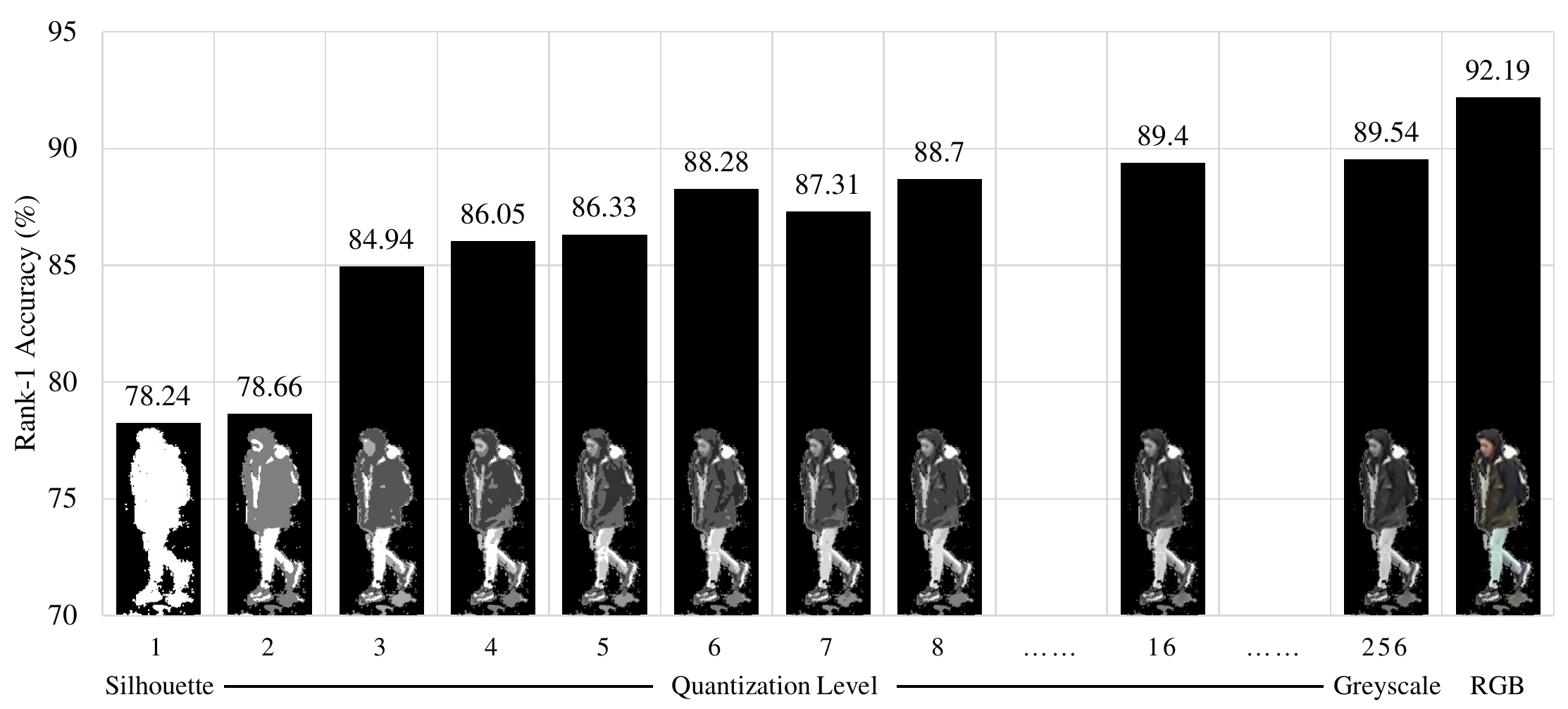}
		\caption{Comparison of the performance with different input images: Silhouettes, grayscale images with various quantization levels and full color images.}
		\label{fig:quantization}
	\end{figure*}
	
	\subsubsection{Impact of Color and Background}
	
	It is interesting that gait recognition and video person re-identification make use of the same kind of input data. When a chromatic image is used, it is referred to as person re-identification, while it is called gait recognition if only binary silhouettes are utilized. Therefore, gait recognition can be viewed as a simplified version of person re-identification. However, compared with the clothing appearance, gait is a much more reliable cue for person authentication. People can change their clothes and issues like workwear or uniform can reduce clothing diversity, which is the basis of person re-identification. Gait is considered stable if the walking speed is normal.
	
	Specifically, in terms of data format, the difference between person re-identification and gait recognition is just the colors in the target set. Therefore, we conduct another experiment where we train the person re-identification models with gait silhouettes on our BUAA-Duke-Gait dataset. The results are shown in the last column of Table~\ref{table:resultreid}. The rank-1 accuracies of these person re-identification models drop dramatically in our dataset. It shows that these models rely heavily on appearance information such as clothing colors of pedestrians for recognition. When we give binary silhouettes to the model rather than the full color images, the model performance degrades from over 95\% rank-1 accuracy to less than 70\%; BiCnet-TKS~\cite{BiCnet-TKS}, achieves less than 15\% rank-1 accuracy. 
	
	We also notice that the background is another influencing factor since irrespective of person re-identification or gait recognition, the pedestrians are represented by rectangular bounding boxes in which the background is inevitably included. To make a comprehensive investigation, we compare four combinations with and without background color and pedestrian color, respectively, in the BUAA-Duke-Gait dataset. The comparison is implemented by changing the input format of our model. The results are shown in Table~\ref{table_imagecolors_Duke-Gait}. Our method can achieve 94\% rank-1 accuracy if the full color images are used. In this case, the task is the same as person re-identification. The current best result on the original DukeMTMC-VideoReID is around 97\%. It shows that even when a person re-identification method is used, our approach is still competitive. The comparison between gait recognition and person re-identification is quantitatively reliable.
	
	\begin{table}[t]
		\setlength{\abovecaptionskip}{1pt}
		\renewcommand{\arraystretch}{1.3}
		\caption{Average rank-1 accuracies of the proposed RealGait in multi-scene recognition on BUAA-Duke-Gait with various combinations of pedestrian and backgrounds color. }
		\setlength{\tabcolsep}{4.5mm}
		\centering
		\begin{tabular}{c|c|c|c}
			\hline
			\multicolumn{2}{c|}{Image}                         & \multicolumn{2}{c}{Background} \\ \cline{3-4} 
			\multicolumn{2}{c|}{Accuracy}              & RGB            &   Subtracted  \\ \hline
			
			\multirow{6}{*}[-37pt]{\rotatebox{90}{Pedestrian}} & \multirow{2}{*}[-25pt]{\rotatebox{90}{RGB}}   & \begin{minipage}[b]{0.2\columnwidth}\centering\raisebox{-1\height}{\includegraphics[width=\linewidth]{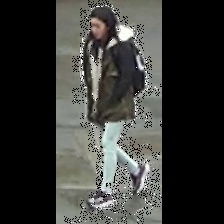}}\end{minipage}  & \begin{minipage}[b]{0.2\columnwidth}\centering\raisebox{-1\height}{\includegraphics[width=\linewidth]{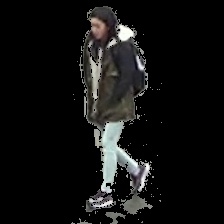}}\end{minipage} \\
			&                        & 94.00\%        & 92.19\%        \\ \cline{2-4}

			& \multirow{2}{*}[-20pt]{\rotatebox{90}{Silhouette}} &  \begin{minipage}[b]{0.2\columnwidth}\centering\raisebox{-1\height}{\includegraphics[width=\linewidth]{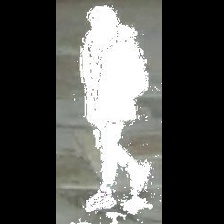}}\end{minipage} & \begin{minipage}[b]{0.2\columnwidth}\centering\raisebox{-1\height}{\includegraphics[width=\linewidth]{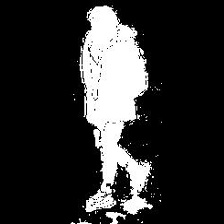}}\end{minipage}
			\\
			&                        & 88.98\%        & 78.24\%        \\ \hline
		\end{tabular}
		\label{table_imagecolors_Duke-Gait}
	\end{table}
	
	If the background color is removed, the performance drops from 94\% to 92.19\%. As can be seen, the performance is still good, which implies that the appearances, including both contour and color, are discriminative. When the color information of pedestrians is further removed, the task is similar to binary silhouette-based gait recognition, and the performance declines to 78.24\%. Although the accuracy is lower than full color identification, it is remarkable that such a simple feature can still achieve acceptable accuracy. These results imply that even in such a challenging task, gait is still a reliable cue for recognition.
	
	It is interesting that when the input is silhouette plus background color, the accuracy can be improved to 88.98\%. By removing the background, the performance differences are 1.81\% vs. 10.6\% for chromatic and binary pedestrian recognition, respectively. We believe that the background color also contains information for pedestrian recognition. As to the difference caused by pedestrian color, it is 5.02\% vs. 13.81\% without background. We can conclude that the pedestrian color has a greater impact than that of the background color. 
	
	To simplify the narrative, we use the terms ``color'' or ``appearance'' in above paragraphs. However, it actually includes two parts: the intensity of illumination which corresponds to the ``V'' channel in the hue, saturation, value (HSV) color space, and the color that corresponds to the ``H'' and ``S'' channels. To further investigate the impact of intensity, we conduct another experiment by comparing different quantization levels. The results are shown in Figure~\ref{fig:quantization}. To generate a gait image with quantization level $m$, $m$-bins quantization scheme is applied to the grayscale pedestrian image. The rank-1 accuracy increases from 78.24\% to 84.94\% with three quantization levels. In other words, just three bins of a grayscale image can bring enough appearance information for recognition. The recognition accuracy gradually increases to around 89\% as the quantization levels rise, which has a performance gap of around 3\% with the full color images.
	
	\begin{figure}[t]
		\centering
		\includegraphics[width=\linewidth]{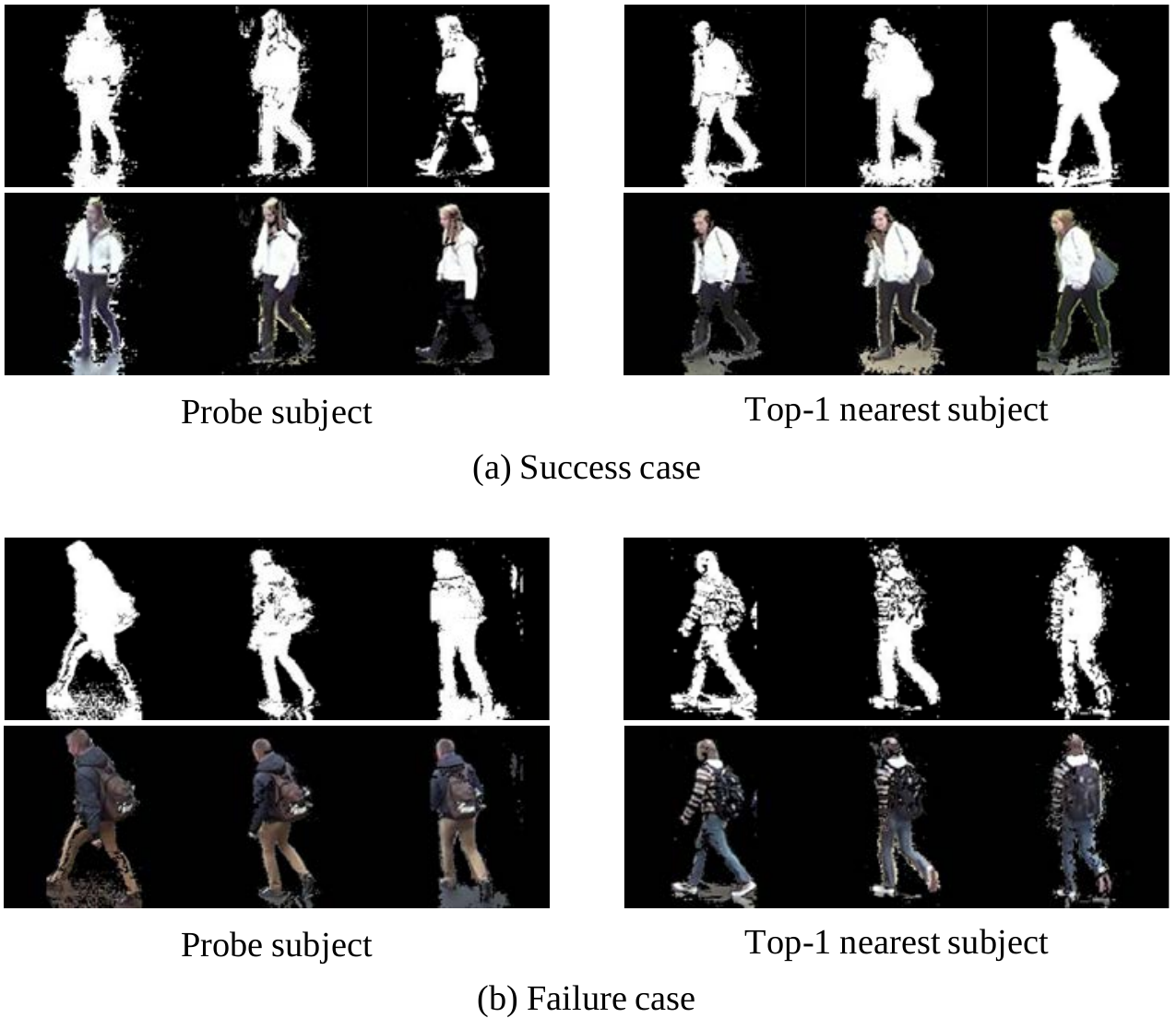}
		\caption{Examples of (a) successful and (b) failed recognition. Top row: binary silhouettes. Bottom row: full color images. Errors may occur in a gait recognition system when body silhouette is severely corrupted or influenced by identity-irrelevant factors.}
		\label{fig:cases}
	\end{figure}
	
	\subsection{Discussion}
	This paper delivers two key messages. The first one is that existing gait recognition approaches cannot meet the requirements of less-controlled scenarios. The merit of both gait and face recognition is that they do not need active user cooperation. Gait is considered to have a much longer effective standoff or range than face. Therefore, the key advantage of gait over other biometric traits is its feasibility in dealing with difficult scenarios. Obviously, current gait recognition research moves in the opposite direction since most studies are based on well-controlled data. However, it does not mean such studies are not useful. In well-controlled scenarios like access control and narrow corridors, conventional gait recognition can achieve very high performance. There are pros and cons to both kinds of scenarios.
	
	Another key message we deliver in this paper is that gait features are probably the basis of a general pedestrian identification task. It should be pointed out that all the four scenarios shown in Table~\ref{table_imagecolors_Duke-Gait} include gait information. Our results contradict the belief that appearance alone is better than gait. Indeed gait plus appearance seems to be better than gait alone, which is reasonable. However, the fact revealed in this paper implies that the gait feature alone is sufficiently reliable for person identification. From an association task within a restricted scenario to open set recognition of more than one thousand people, person re-identification achieved good success that cannot be simply explained by clothing diversity. The ablation study in this paper shows that gait is probably the true reason behind the success of re-identification.
	
	To better demonstrate the subtle difference between gait recognition and appearance-based person re-identification, we show two examples in Figure~\ref{fig:cases}, including a successful recognition case and a failure case. As can be seen, gait features can deal with large view differences, which is considered difficult for appearance-based recognition. However, errors may occur when the body silhouette is severely corrupted or influenced by identity-irrelevant factors, such as a big backpack. In this case (top row of Figure~\ref{fig:cases} (b)), even for humans, it is difficult to tell the difference. However, if color information is available (bottom row of Figure~\ref{fig:cases} (b)), the task becomes much easier. Although gait is reliable biometrics, it does have certain limitations.

	\section{Conclusion}
	\label{sec:conclude}
	
	In this paper, we consider the image-based gait recognition problem in a real surveillance scenario. Since existing gait datasets are captured in controlled environments, factors like background clutter, presence of other people in the video, unexpected walking direction, and complex walking patterns are ruled out. Consequently, models developed from such datasets do not perform well for gait recognition in real-world applications. To this end, we construct a new gait dataset named BUAA-Duke-Gait by extracting silhouettes from an existing video person re-identification dataset. It is not only a new challenging benchmark for gait recognition but also provides a benchmark for comparing gait recognition and person re-identification performance. 
	
	A novel gait recognition method is proposed to meet the challenges posed in surveillance videos. Experimental results on the new dataset show that the proposed method outperforms the state-of-the-art in gait recognition. We also compare the proposed binary silhouette-based gait recognition method with the full color person re-identification task. Our method achieves a rank-1 accuracy of 78.24\% in the silhouette-based gait recognition task and 94\% in the person re-identification task. These results show that recognizing people by gait in real surveillance is feasible, and gait plays an important role in video person re-identification.
	
	
	\bibliographystyle{IEEEtran}
	\bibliography{IEEEabrv,ref}
	
	\newpage

	

\end{document}